\pdfoutput=1

\documentclass[acmsmall,xcolor=table, manuscript,screen]{acmart}
\AtBeginDocument{%
  }

\usepackage{adjustbox}
\usepackage{multirow}
\usepackage{enumitem}
\usepackage{colortbl}

\def\tsc#1{\csdef{#1}{\textsc{\lowercase{#1}}\xspace}}
\tsc{WGM}
\tsc{QE}

\newcommand*{\rowstyle}[1]{
  \gdef\@rowstyle{#1}%
  \@rowstyle\ignorespaces%
}

\newcolumntype{=}{
  >{\gdef\@rowstyle{}}%
}

\newcolumntype{+}{
  >{\@rowstyle}%
}

\makeatother

\title{Impact of Visual Context on Noisy Multimodal NMT: An Empirical Study for English to Indian Languages}

\author{Baban Gain}
\orcid{0000-0001-8673-7078}
\affiliation{%
  \institution{Indian Institute of Technology Patna} \country{India}
}
\email{gainbaban@gmail.com}

\author{Dibyanayan Bandyopadhyay}
\orcid{https://orcid.org/0000-0001-5279-6344}
\affiliation{%
  \institution{Indian Institute of Technology Patna} \country{India}
}

\author{Samrat Mukherjee}
\orcid{0009-0009-9528-6681}
\affiliation{%
  \institution{Indian Institute of Technology Patna} \country{India}
}

\author{Chandranath Adak}
\orcid{0000-0002-9085-2770}
\affiliation{%
  \institution{Indian Institute of Technology Patna} \country{India}
}
\email{chandranath@iitp.ac.in}

\author{Asif Ekbal}
\orcid{0000-0003-3612-8834}
\affiliation{%
  \institution{Indian Institute of Technology Jodhpur \& Indian Institute of Technology Patna}
  \country{India}
}
\email{asif@iitj.ac.in}

\renewcommand{\shortauthors}{Gain et al.}

\begin{abstract}
Neural Machine Translation (NMT) has made remarkable progress using large-scale textual data, but the potential of incorporating multimodal inputs, especially visual information, remains underexplored in high-resource settings. While prior research has focused on using multimodal data in low-resource scenarios, this study examines how image features impact translation when added to a large-scale, pre-trained unimodal NMT system. Surprisingly, the study finds that images might be redundant in this context. Additionally, the research introduces synthetic noise to assess whether images help the model handle textual noise. Multimodal models slightly outperform text-only models in noisy settings, even when random images are used. The study's experiments translate from English to Hindi, Bengali, and Malayalam, significantly outperforming state-of-the-art benchmarks. Interestingly, the effect of visual context varies with the level of source text noise: no visual context works best for non-noisy translations, cropped image features are optimal for low noise, and full image features perform better in high-noise scenarios. This sheds light on the role of visual context, especially in noisy settings, and opens up a new research direction for Noisy Neural Machine Translation in multimodal setups. The research emphasizes the importance of combining visual and textual information to improve translation across various environments. 
Our code is publicly available at {\url{https://github.com/babangain/indicMMT}}.

\end{abstract}



\begin{document}

\maketitle

\section{Introduction}

Unimodal Neural Machine Translation (UNMT) systems generate reliable outputs across many domains and languages. However, these models often fail to translate ambiguous words that require additional contextual information for disambiguation. This issue is further amplified when the dataset is noisy, containing spelling or grammatical errors. Additionally, UNMT models struggle to produce context-dependent translations. Multimodal Neural Machine Translation (MMT) is particularly important in the era of social media.

In MMT, an image is used as contextual information to support the translation process. For example, in the sentence \emph{"The color of the court is red,"} the word \emph{court} can be interpreted in multiple ways. One interpretation is a playground for indoor games like tennis or badminton, while another could refer to a judicial court. Although several techniques have been proposed for MMT, most have been applied to high-resource language pairs, such as English--German or English--French. These languages belong to the same linguistic family, use the Latin script, and often share common vocabulary. Translating between distant language pairs, e.g., English and Hindi, is more challenging, as they share fewer commonalities. 
Furthermore, Hindi sentences are gender-dependent, whereas English is not~\cite{comrie2009world}. For example, the translation of \emph{"I am going to the market."} differs depending on whether the subject is male or female. Although there have been a few efforts to incorporate multimodal information for translating English to Hindi, most do not leverage large-scale unimodal pre-training. Existing methods either omit large-scale unimodal data entirely or use it only for limited purposes such as word embedding training. While multimodal translation is inherently different from unimodal translation, training encoders and decoders with large-scale unimodal data may help achieve better translation performance. 
Although object tags can aid translation quality~\cite{gupta-etal-2021-vita}, relying solely on object tags without incorporating additional visual information from the image may be suboptimal for the following reasons:

\begin{figure*}
 \centering
 \includegraphics[width=0.8\textwidth,keepaspectratio]{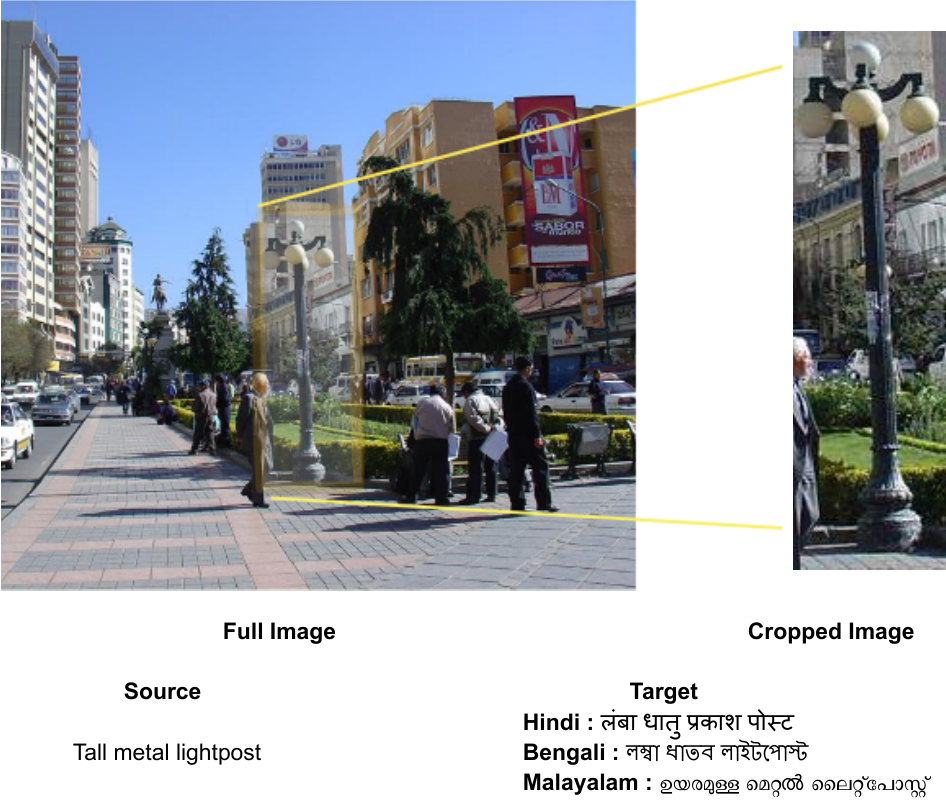}
 \caption{Example of combined Hindi, Bengali, and Malayalam dataset}
 \label{fig:dataset_example}
\end{figure*}
\begin{itemize}[leftmargin=*]

    

\item Only object tags (such as \emph{person, ball, tree}, etc.) are linked to the source text; other details about the objects and images that could assist in translation are not utilized.

\item If there is no additional information about the objects, their locations, or their interactions, the model may use object tags misleadingly. For instance, the model may fail to determine whether the subject is \emph{male} or \emph{female} if another \emph{male} person is present in the image and the subject is actually \emph{female}, as both would appear in the list of detected objects regardless of their actions.

\item For each detected object, only one tag with the highest confidence is selected. Misclassified tags can mislead the NMT model because tag selection is discrete, and the object detection model is prone to errors.

\item Due to inadequate training of the object detection model, many objects may remain unclassified.

\end{itemize}
The aforementioned factors prevented object tag-based models from consistently improving the results.
Therefore, it is preferable to use and incorporate visual features that have been extracted from pre-trained image models, e.g., Vision Transformer (ViT), VGG, ResNet \citep{he2015deep}, etc. 
When source texts are hidden, some probing tasks demonstrate how MMT models can retrieve missing data, such as color, person, etc., from visual features \cite{li-etal-2022-vision}. Here, some part of the textual input is masked, and the model is trained to translate the sentence with the help of the image, forcing it to focus more on the visual features in order to retrieve information about masked input tokens. However, it may not be a realistic strategy in scenarios involving incomplete information. Any mask token will not be present in real-world datasets, where some of the information is missing from the source text, and it may be challenging to infer that the information is insufficient.
In summary, efforts have been made to make use of visual features in the MT system directly. Most of them only use multimodal data, which is scarcely available. Additionally, other techniques that rely on strong baselines only use image object tags when fine-tuning. 
In this study, we investigate the circumstances in which visual features have a positive impact on MMT systems. We use large pre-trained models as a baseline so that the models are robust enough to distinguish words from intra-sentence context when available. We observe that images have a negligible contribution in non-noisy settings. We further observe that the models are able to utilize cropped image features in low-noisy settings, and full image features are more effective in high-noisy settings. Our study indicates that visual features' impact on the model performance increases with noise in the source text. 

The major \textbf{contributions} of this paper are briefly outlined below.

\textbf{\em (i) Noisy MMT:}
We develop our systems for noisy MMT. To the best of our knowledge, this is the earliest attempt of its kind to translate noisy utterances in multimodal settings. We observe an increased positive impact of visual context in the presence of noise in the dataset.

\textbf{\em (ii) Multilingual MMT:} 
We perform multilingual multimodal fine-tuning and utilize a multilingual pre-trained model. 
This is the first attempt at a multilingual setting that includes Indic languages.

\textbf{\em (iii) Probing the Need for Images:} 
By substituting relevant images corresponding to the text with random ones, we investigate the effectiveness of using semantically aligned visual information. This is the first attempt of its kind to conduct such probing in Indian languages.

\textbf{\em (iv) Strong UNMT Baseline:} 
We use a strong off-the-shelf UNMT model for fine-tuning. 
This is essential to determine whether the improved performance of multimodal methods stems from the extracted information in images or from the limitations of the text-only model in handling ambiguity.

\textbf{\em (v) Enhanced Visual Features in Indian Settings:} 
We extract rich visual features from ViT as well as CLIP instead of using CNN-based features adopted in contemporary Indic MMT methods.

\textbf{\em (vi) State-of-the-Art Results:} This work primarily aims to understand the impact of visual features on multimodal English-to-Indic language translation in the presence of noise. While our goal is not solely to achieve state-of-the-art (SOTA) performance, our model obtains SOTA results for all three target Indic languages (Hindi, Bengali, and Malayalam), with English as the source language.

The rest of the paper is organized as follows. 
In Section \ref{sec:related-works}, we discuss related MMT methods along with models for Indic languages and noisy text translation. 
Thereafter, Section \ref{sec:datasets} explains the employed dataset and the noise addition procedure. 
Subsequently, Section \ref{sec:methodology} mentions our methodology, and Section \ref{sec:results} and Section \ref{sec:ablation} discuss the results with observations and ablations. Finally, Section \ref{conclusion} concludes this paper.

\section{Related Works}
\label{sec:related-works}

MMT has seen growing interest due to its ability to incorporate complementary visual context to improve translation quality. However, the progress in this area is constrained by the limited availability of large-scale source-target-image triplet datasets. Despite this, a wide range of approaches have been proposed to address the challenges in MMT through innovations in model architecture, integration of visual features, and adaptations for low-resource and linguistically diverse settings. This section surveys major developments in general MMT, its application to Indian languages, context-aware translation, and robustness to noisy input.

\subsection{Multimodal Translation}
\label{subsection:related-work-mmt}

One of the earliest attempts for MMT was to integrate visual features with shared attention mechanism \citep{caglayan-etal-2016-multimodality}. However, it did not result in performance improvement.
Other early attempts include the concatenation of text features with global and regional visual features \citep{huang-etal-2016-attention} to generate Multimodal Translation. Here, regional features are treated as tokens of the input sentence.
Subsequently, a  multi-task-based training framework \citep{elliott-kadar-2017-imagination} was employed with a shared encoder where the tasks included standard NMT tasks and visual feature reconstruction.
The following attempts to tackle challenges included consideration of visual features as pseudo-text features \citep{calixto-liu-2017-incorporating}. First, they projected visual feature vectors into source text embedding space and used them as the first and/or last word of the source sentence. Further, they use them to initialize the hidden state of the encoder instead of using a zero vector. Finally, they use the projected image representations and encoder output to initialize hidden representations of a decoder. They use bidirectional RNN as the encoder. 
Since visual features are needed in very specific cases, the model tends to learn to ignore visual features, resulting in ignoring them, even when necessary. Therefore, a visual attention grounding mechanism \citep{zhou-etal-2018-visual} was adopted for MMT  that links the visual semantics with the corresponding textual semantics.

In order to enable the MMT systems to adequately handle multimodal information, a translate and refine approach \citep{ive-etal-2019-distilling} was adopted for MMT, where the first stage decoder generates translation on textual information and second stage decoder refines it with the help of the visual features.
To improve the interaction between textual and visual features, a latent variable was added to model \citep{calixto-etal-2019-latent}. This latent variable can be seen as a stochastic embedding that is used by the decoder for translation, as well as it can be used to predict image features.
Attention \citep{attention-is-all-you-need} based multimodal interactions are widely used in MMT. This includes cross-attention between regional image features and decoder self-attention output \citep{ZHAO20221-region-attentive}.

Multimodal information generally helped in obtaining better results. However, it was observed that many of the improvements were not related to the presence of visual features in MMT. This indicated a need to probe how and why MMT models generate better translation compared to UNMT models.
In a probing task \citep{li-etal-2022-vision}, the words that included color information (red, green, etc.) were masked. In addition, words that indicate characters and nouns were also replaced with mask tokens from the source text to simulate a situation where information is insufficient in the source text. They proposed a Selective Attention-based transformer that is able to recover the masked information from the image at the generated translation better than other existing models.
Similarly, it was observed that models can integrate the visual modality if they are complementary rather than redundant \citep{caglayan-etal-2019-probing}. Further, they observed that visual features increase the robustness of NMT by acting as input noise.

\subsection{Multimodal Translation on Indian Languages}
\label{subsection:related-work-indian}

The early attempts for MMT for Indian languages relied on synthetic data
\citep{dutta-chowdhury-etal-2018-multimodal}, which was generated by using a text-based PBSMT model. Then, they use the projected image feature for hidden representations of encoder and decoder \citep{calixto-liu-2017-incorporating} to generate Multimodal MT. Initializing the hidden representation of the decoder generated better results compared to that of the encoder.
Subsequently, it was observed that using multimodal methods achieves better results \citep{sanayai-meetei-etal-2019-wat2019,laskar-etal-2019-english} compared to UNMT or image captioning.
Doubly Attentive decoder \citep{calixto-etal-2017-doubly} architecture, which was proven to generate a better translation for other languages, was adopted for English-Hindi \citep{laskar-etal-2020-multimodal,laskar-etal-2021-improved}. They use pre-trained word embedding learned from IITB Corpus \citep{kunchukuttan-etal-2018-iit} to initialize the models and augment datasets by using Giza++ Tool \citep{och03:asc-giza-plus-plus}, to obtain phrase pairs. Further, the pre-trained VGG19 \citep{https://doi.org/10.48550/arxiv.1409.1556-vgg} model is used to extract image features.

It was observed that using text-based pre-trained models achieved better results for Hindi VG Test set compared to the multimodal model trained only on multimodal dataset \citep{gain-etal-2021-iitp}. However, it obtained lower results for the Challenge set. Further, the text-based model was trained on a larger bilingual corpus, which could be a major contributing factor to better results. Multimodal Transformer architecture was adopted for English-Hindi \citep{shi2022adding-visual-info}, which improved the results compared to its UNMT baseline.
There are other works that utilized multiple captions from the dataset to train the models \citep{singh-etal-2021-multiple}. It was observed that multiple captions introduce lexical diversity, which helps to generate robust translation.

Only a limited number of studies have employed large-scale pre-training. 
A model was pre-trained on IITB corpus \citep{kunchukuttan-etal-2018-iit}, and it was observed that large-scale pre-training is more important for better translation in contrast to image features \citep{parida-etal-2019-idiap}. In another study, the multimodal features are found to be less relevant as a good pre-trained model can handle ambiguity issues \citep{10.1007/978-981-19-4831-2_5-caption-translation-object}.
A pre-trained mBART model was used to initialize the weights \citep{gupta-etal-2021-vita}. Then, they used unimodal data for the first stage of fine-tuning. Finally, they concatenated object tags with source text before last-stage fine-tuning. Those methods are entirely unimodal except for the usage of Object detection models for object tag extraction at data pre-processing stages. Thereafter, a similar method was adopted for English-Bengali \citep{parida-etal-2021-multimodal}. Inspired from \cite{gronroos-etal-2018-memad}, a sequence-to-sequence model was used for MMT \citep{parida-etal-2022-silo}. In this method, the input is the concatenated source and target side, and the output is object tags,  which act as dummy image features on unimodal datasets. They follow the object tag concatenation-based method, which resulted in huge BLEU gains compared to the text-only model. However, the text-only models were not trained on very large corpus in required languages. This makes them unable to disambiguate sentences properly.

There were some attempts in the reverse direction, where Spatial attention was used for image features \citep{MEETEI20232102-hindi-english-news} for Hindi to English Multimodal News Translation. 
In the meantime, some MMT methods were proposed for other low-resource Indian languages. A transliteration-based phrase augmentation method was adopted for English-Assamese \citep{LASKAR2023979-english-assamese-mmt-transliteration} with Assamese VisualGenome dataset \citep{9752181-mmt-for-english-assaemse} and English-Bengali \citep{laskar-etal-2022-english}. A similar method was used for English-Hindi \citep{laskar-etal-2022-investigation-english}, and this method was able to generate better results compared to UNMT. A comparison of results was performed on the Mizo VisualGenome dataset \citep{mizo-vis-genome} with Transformer, BiLSTM, and PBSMT models \citep{10.1145/3445974-mizo-multimodal-improved, Pathak2018EnglishMizoMT-english-mizo-neural-statistical,9500022-mt-english-to-mizo}, where the transformer outperforms the other models. However, it was observed that NMT-based models generate a poor-quality translation for longer sentences. A multilingual multimodal NMT system was introduced for English and Dravidian languages \citep{chakravarthi-etal-2019-multilingual}, which generated better results compared to UNMT and single-language MMT. All these methods assumed the presence of clean and grammatical training data, which may not be very suitable for real-world use cases.
\subsection{Context-Aware Translation}
Apart from using multimodal information as context, there have been several attempts to use unimodal information as a context. For example, the previous and next sentences can serve as a useful context for document translation. 
Similarly, previous sentences in a conversation can serve as a context for chat translation. The context is generally incorporated by concatenation \cite{tiedemann-scherrer-2017-neural,gain-et-al-not-all} or using a different encoder for context. Hierarchical Attention Networks can achieve better results \cite{miculicich-etal-2018-document} compared to context-agnostic NMT. It was observed that performance gains from multi-encoder models cannot be entirely attributed to the contextual information. Setting the weight of the context to zero during testing does not have a significant negative effect on the results\cite{li-etal-2020-multi-encoder,gain-etal-2022-investigating}. Similarly, training and testing with random context also improve the results \cite{appicharla2023case}. The standard metrics for NMT were unable to capture the improvements at the discourse level. \cite{fernandes-etal-2023-translation} proposed a discourse-aware framework for performance evaluation.
\subsection{Noisy Neural Machine Translation}
The presence of noise in the data has adverse effects on the translation quality \citep{belinkov2018synthetic,khayrallah-koehn-2018-impact}. Early attempts for Noisy MT relied on Normalization of text \citep{wang-ng-2013-beam}.
Due to the lack of parallel corpora for Noisy NMT, most of the research relies on synthetically generated noise. This method helped to achieve better results in translating erroneous speech transcripts \citep{xie2017data-noising,sperber-etal-2017-toward}. Subsequently, the noise was added with texts for different datasets with multiple types of synthetic noise \citep{karpukhin-etal-2019-training, cheng-etal-2018-towards}.
Generally, noisy text translation can be viewed as a domain adaptation task. Fine-tuning on noisy data improves the robustness of the model \citep{helcl-etal-2019-cuni,berard-etal-2019-naver,dabre-sumita-2019-nicts-supervised}. Other methods include the construction of source and target side adversarial examples \citep{cheng-etal-2019-robust} for noisy NMT.
Subsequently, Insertion and Deletion predictors were added to the NMT, which improved results on noisy inputs \citep{wang-etal-2021-secoco-self}. It was observed that multi-tasking with translation and denoising objectives helps to generate robust translation \citep{passban-etal-2021-revisiting-robust}. Further, they observed improvements after adding synthetic noise to the dataset and using Dual Channel Decoding. Noisy text translation is an unexplored area in Indian languages. It was observed that fine-tuning with noisy data improves the results for Conversational NMT \citep{gain-etal-2022-low}. However, since the training datasets were synthetically generated, the improvements are mostly due to domain adaptation.

\section{Dataset Details}
\label{sec:datasets}

 
 
 

We use the VisualGenome (VG) datasets containing Bengali~\citep{sen2022bengali-visgen}, Hindi~\citep{hindi-visual-genome:2019}, and Malayalam~\citep{11234/1-3533-malayalam-visgen} sentences for fine-tuning. The train, validation, test, and challenge subsets of each dataset contain 29230, 998, 1595, and 1400 sentences, respectively. The challenge subset comprises sentences with potential ambiguity (such as “court”) that may or may not be resolved using textual information alone. 
Each row in a dataset contains the following fields:
\begin{itemize}[leftmargin=*]
  \item \textbf{\em Source text:} The utterance in English.
  \item \textbf{\em Target text:} The corresponding utterance in Hindi, Bengali, or Malayalam.
  \item \textbf{\em Image ID:} The filename of the image relevant to the utterance.
  \item \textbf{\em Coordinates of image:} The dataset provides the bounding box of the specific image region on which the utterance is based.
\end{itemize}

\begin{table}[!htb]
\caption{Statistics of Hin: Hindi, Beng: Bengali, and Mala: Malayalam VisualGenome datasets}
\centering
\resizebox{0.6\textwidth}{!}{
\begin{tabular}{c|c|c|c|c|c}
\hline
\multirow{2}{*}{\textbf{Subset}} & \multirow{2}{*}{\textbf{\#Sentences}} & \multicolumn{4}{c}{\textbf{\#Average Words per Utterance}} \\ \cline{3-6} 
 & & \textbf{English} & \textbf{~Hindi~} & \textbf{~Bengali~} & \textbf{Malayalam} \\ 
 \hline \hline
\textit{Train} & 28930 & 4.95 & 5.03 & 3.94 & 3.70 \\
\textit{Valid} & 998 & 4.93 & 4.99 & 3.94 & 3.63 \\
\textit{Test} & 1595 & 4.92 & 4.92 & 4.02 & 3.57 \\
\textit{Challenge} & 1400 & 5.85 & 6.17 & 4.76 & 4.32 \\ \hline
\end{tabular}}
\label{tab:visgenome-stats}
\end{table}


A sample of the dataset is presented in \autoref{fig:dataset_example}, and we describe some additional dataset statistics in \autoref{tab:visgenome-stats}.
We frequently noticed that the image size within the box's boundaries is insufficient. More information might be available if the entire image were used rather than just the cropped area contained within the bounding box. A study \cite{gain-etal-2021-experiences} observed that using the cropped image feature of ResNet-50 yields better results compared to the full image. However, the difference is marginal in most experimental settings, and using full images achieved slightly better results in one setting. Therefore, we perform our experiments in both setups.

\subsection{Noisy Data Generation}
We generate noise in the datasets as they appear in social media. Social media captions frequently misspell words with two identical consecutive characters, write words without vowels, and omit articles (a, an, the). Since our work intends to find if the visual context can help to translate noisy text, we did not add noise to the images.

\subsubsection{Low Noise}
We iterate through all the words in a sentence and remove them if we encounter an article with 0.2 probability. If the word is non-article or article but not removed in the previous step, we look for vowel characters in the word and remove them with 0.1 probability. Finally, we check if the word contains the same letters at two consecutive positions. In that case, we remove one of them with 0.2 probability. 
We started with 0.3 probability for each noise generation operation and randomly selected 20 samples from the dataset. 
We engaged some human annotators who are active on social media to rate noised sentences in terms of naturalness from 1 to 5. A rating of 5 indicates that these types of noisy sentences often appear in social media, and a rating of 1 indicates that the appearance of these types of noisy sentences is quite unlikely. We keep reducing the probability of each type of noise by 0.1 until we reach at least a 4.5 rating on average. 

The statistics of added noise with respect to the actual sentences are reported in \autoref{tab:stat-noise}. As we acquire the noisy version of the data through subjective human judgment, we use the BLEU \citep{papineni-etal-2002-bleu} metric to report n-gram matches as a measure of similarity.
Similarly, TER \citep{snover-etal-2006-study} indicates the number of edits required to obtain the original source from the corrupted text. chrF2 \citep{popovic-2015-chrf,popovic-2016-chrf} is a character-level F-score metric. To maintain a consistent distribution of noise throughout training, validation, and evaluation, we introduce noise to all the available subsets of the data. In low noise settings, the resulting BLEU scores typically fall within the range of 65.9 to 67.5, while chrF2 scores range from 87.0 to 87.7; a lower TER score, in this case, indicates a lower amount of noise, which ranges from 16.2 to 16.5 in low noise settings.


\begin{table}[]
\caption{Statistics of the source text after addition of noise when compared with original source text }
\label{tab:stat-noise}
\resizebox{0.6\textwidth}{!}{
\begin{tabular}{l|lll|lll}
\hline
\multirow{2}{*}{\textbf{Dataset}} & \multicolumn{3}{c|}{\textbf{Low Noise}} & \multicolumn{3}{c}{\textbf{High Noise}} \\ \cline{2-7} 
& \multicolumn{1}{c|}{\textbf{BLEU}} & \multicolumn{1}{c|}{\textbf{chrF2}} & \multicolumn{1}{c|}{\textbf{TER}} & \multicolumn{1}{c|}{\textbf{BLEU}} & \multicolumn{1}{c|}{\textbf{chrF2}} & \multicolumn{1}{c}{\textbf{TER}} \\ 
\hline \hline
\textit{Train} & \multicolumn{1}{l|}{67.5} & \multicolumn{1}{l|}{87.3} & 16.2 & \multicolumn{1}{l|}{38.1} & \multicolumn{1}{l|}{71.1} & 34.9 \\ 
\textit{Valid} & \multicolumn{1}{l|}{66.9} & \multicolumn{1}{l|}{87.1} & 16.4 & \multicolumn{1}{l|}{36.7} & \multicolumn{1}{l|}{70.6} & 35.5 \\ 
\textit{Test} & \multicolumn{1}{l|}{67.0} & \multicolumn{1}{l|}{87.0} & 16.5 & \multicolumn{1}{l|}{36.9} & \multicolumn{1}{l|}{70.6} & 35.3 \\ 
\textit{Challenge} & \multicolumn{1}{l|}{65.9} & \multicolumn{1}{l|}{87.7} & 16.5 & \multicolumn{1}{l|}{37.3} & \multicolumn{1}{l|}{71.9} & 36.0 \\ \hline

\end{tabular}}
\end{table}
\subsubsection{High Noise}
We remove articles with 0.3 probability. Unlike low noise, we do not make further modifications to articles if they are not removed because we are already adding other noises in high amount. Then, we remove vowels and repeat characters with 0.3 probability each. 
In scenarios with high noise levels, the achieved BLEU scores usually lie between 36.7 to 38.1, and the corresponding chrF2 scores range from 70.6 to 71.9. A lower TER score in such cases suggests a reduced amount of noise, which typically falls within the range of 34.9 to 36.0 in high-noise settings.

\subsection{Pre-Processing}
\label{sub:preprocessing}
We use langid.py \citep{lui-baldwin-2012-langid}, a language identification tool to detect if any sentence pair contains the wrong language.
We found some alignment issues in the corpus, such as the presence of source sentences in the target field of the dataset.
Some human annotators manually annotated those instances by translating them into the target language, taking the corresponding image into consideration. Further, the source data was unavailable in a few instances of Hindi VisualGenome. Since the Bengali VisualGenome is aligned with Hindi VisualGenome, we copy the corresponding source sentence from Bengali VisualGenome. Finally, while Malayalam VisualGenome contains the same set of sentences, image pairs are Hindi and Bengali VG; they are not aligned. We align them to match the order of Bengali VG. Aligning the datasets helps to reduce memory usage, as visual features can be common between all languages. Finally, we follow the same pre-processing steps that are performed to train the pre-trained model. Since Hindi, Bengali, and Malayalam are written in different scripts, we convert all target side data to the Devanagari script by a publicly available tool \citep{kunchukuttan2020indicnlp}.
We use Hindi, Bengali, and Malayalam VisualGenome datasets to fine-tune all our models. 
For cropped experiments, we crop image portions within bounding boxes. We extract (cropped as well as full) image features from vit\_base\_patch16\_224 model \citep{50650-vision-transformer} unless mentioned otherwise.
\section{Methodology}
\label{sec:methodology}

In the literature, previous probing methods focused on the usability of images when some information from the source was deprived of the model. 
For example, the words with \emph{color, person}, etc., which can be available in the image, are masked. Although the models function well in these circumstances, masking words in the source sentence explicitly tells the model that some information is hidden. This forces the model to look for pertinent data in the image. However, in most situations, there will not be explicit mask tokens that indicate that a word is absent. On the other hand, simply dropping the words (without mask token) and expecting the models to retrieve them does not align with the MT objective. 
For example, in the source sentence \textit{chinese characters on a red vehicle}, if we remove the word \textit{red}, then \textit{chinese characters on a vehicle} is also a valid sentence, and the model should not attempt to add information related to color in the generated translation. Doing this would lead to undesirable information in the generated sentence, which is absent in the source.
Therefore, instead of providing any information if the input sentence is noisy, we try to determine whether the models can use information from images in the presence of noise in the source text. 

\subsection{Baseline}

\begin{figure*}
    \centering
    \includegraphics[width=\textwidth]{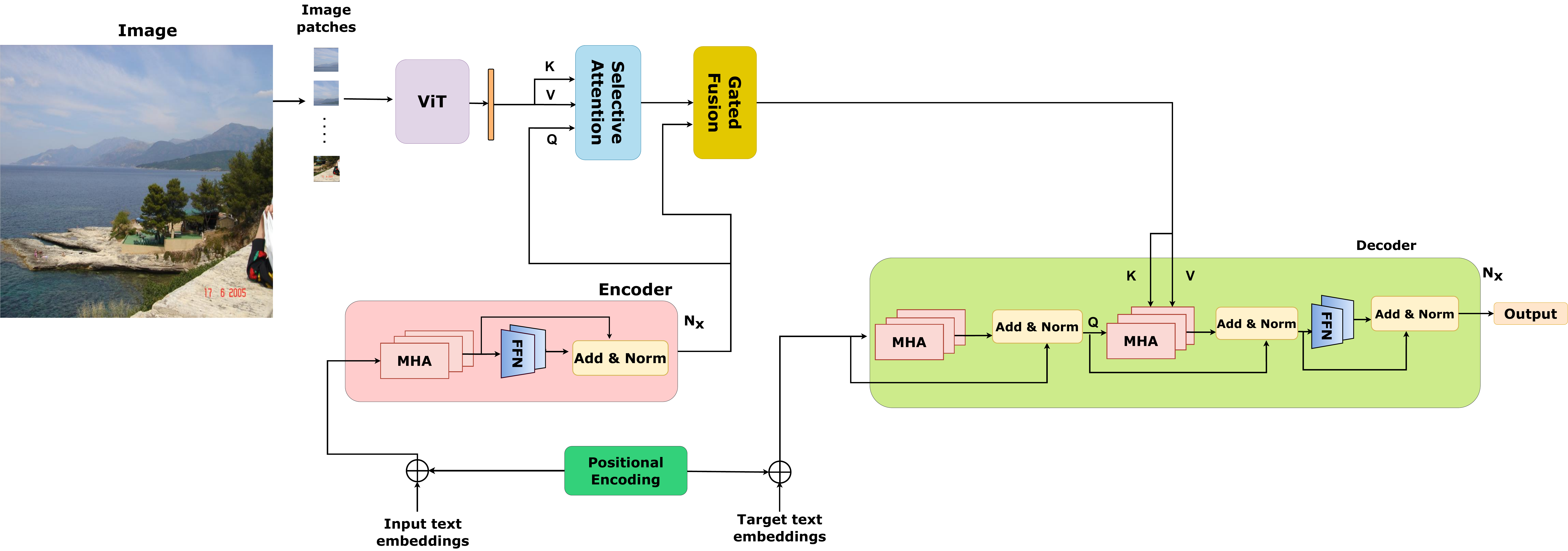}
    \caption{Selective Attention Architecture for Multimodal MT. (softcopy after zooming-in exhibits better display)}
    \label{fig:selective_attn}
\end{figure*}

We use a transformer-4x model\footnote{\url{https://ai4b-public-nlu-nlg.objectstore.e2enetworks.net/en2indic.zip}} trained on Samanantar \citep{ramesh-etal-2022-samanantar} corpus containing 49.6 million sentence pairs between English and 11 Indian languages as our baseline. The model consists of six encoder and six decoder layers. We initialize weights from this model for all our experiments and use the same dictionary for byte-pair-encoding. For the English-Malayalam Challenge set, it achieves a 26.3 BLEU score, whereas the SOTA system \citep{parida-etal-2022-silo,gupta-etal-2021-vita} achieves a 20.4 BLEU score. This outperforms SOTA by 5.9 BLEU points even before fine-tuning on any data. After fine-tuning the model on only source-target sentence pairs on respective VisualGenome data, we observe that it outperforms all the existing multimodal models by a large margin. Considering the additional overhead of handling images and extracting visual features, we attempt to find if they are worth it when we can simply use a text-based model only and achieve better results since text-only data is available in large quantities.

\subsection{Unimodal Fine-tuning}
First, we follow pre-processing steps as mentioned in \autoref{sub:preprocessing} and combine data from Hindi, Bengali, and Malayalam VG datasets. Then, we initialize the encoder and decoder weights from the baseline model (including word embedding) and continue training. This model surpasses all existing baselines on all the languages and both Test and Challenge subsets by large margins. This model does not use any information from images. Unimodal data are widely available, whereas multimodal datasets only have a small number of examples. If multimodal models are not trained on unimodal data, it is difficult to produce reliable outputs. 

\subsection{Multimodal Fine-tuning with Selective Attention}\

For Multimodal Translation, we follow the Selective Attention architecture as introduced in \citep{li-etal-2022-vision}. 
The architecture of the model has been shown in \autoref{fig:selective_attn}. For a given data point of <image, source sentence, target sentence> in the training dataset, at first, we obtain the word and positional embedding for the given source text. The Positional encoding describes the location or position of a token in a sequence in such a way that each position is assigned a unique representation. In Transformers positional encoding, each position is mapped to a vector. As an example, for an input sequence of length L, the positional encoding of the $k$th object within this sequence is given by sine and cosine functions of varying frequencies as follows:
\begin{equation}
    P(k, 2i) = {sin}\left(\frac{k}{n^{2i/d}}\right) ~;~ 
    P(k, 2i+1) = {cos}\left(\frac{k}{n^{2i/d}}\right) ;  
\end{equation}
where, $n$ is a user-defined scalar, and $i$ is used for mapping the indices of the column.
The obtained representation is forwarded through the encoder layer of the Transformer \citep{NIPS2017_3f5ee243-vaswani-transformer}. 
Each encoder layer consists of three major components: 
Multi-Head Attention (MHA) layers, 
Fully Connected layers and 
the Add \& Norm layer.  
MHA is a module for attention mechanisms that simultaneously runs through multiple attention mechanisms. After that, the independent attention outputs are combined and transformed linearly into the expected dimension. Intuitively, having multiple attention heads makes it possible to focus on different parts of the sequence simultaneously \citep{NIPS2017_3f5ee243-vaswani-transformer}.

After the self-attention mechanism has computed the attention scores between each input token and every other token in the sequence, each token's representation is passed through the feedforward network. 
The output of the feedforward network is then added back to the original input representation of the token, which helps to capture more complex relationships between tokens in the sequence. 
The Add \& Norm layer acts as a residual connection and performs normalization of the outputs at the corresponding sub-layer.

To obtain the representation $H_{img}$ of the corresponding input image, we pass the image through the pre-trained Vision Transformer. It is one of the recent architectures for image classification, which employs a Transformer-like architecture over patches of the image. Any image is first split into fixed-size patches. Each patch is then linearly embedded, after which the position embeddings are added, and the resultant vector is fed to a standard Transformer encoder.

Then, we add an attention layer, where $H_{text}$ is the <query> and $H_{img}$ is used as <key and value> \citep{50650-vision-transformer}.
We perform selective attention on both the $H_{img}$ and $H_{text}$, where $H_{img}$ acts as <key and value> pair and $H_{text}$ acts as a query for the computation of the attention as follows.
\begin{equation}
\label{equation:1}
    	H^{\textrm{attn}}_{img} = {softmax}\left(\frac{H_{text} Q ({H_{img}K)^{\textrm{T}}}}{\sqrt{d_k}}\right)H_{img}V;
\end{equation}
where, $Q$, $K$, and $V$ are the layer-specific query, key, and value vectors. 
Subsequently, text-encoder output $H_{text}$ and selective-attention output $H^{\textrm{attn}}_{img}$ are combined with gated fusion to obtain the final encoder representation $Enc_{out}$.
\begin{equation}
\begin{split}
 Enc_{out} = (1 - \lambda) ~  H_{text} + \lambda ~ H^{attn}_{img} ~;~ 
    \lambda = {sigmoid}\left(W_t{H_{text}} + W_i{H^{attn}_{img}}\right); \label{eq:gate}
\end{split}
\end{equation}
where, $\lambda$ indicates the importance of visual features compared to text features. Here, $W_t$ and $W_i$ are trainable parameters. 

Finally, $Enc_{out}$ is forwarded to the decoder. In addition to the encoder sublayers, the decoder contains an MHA layer for Encoder-Decoder cross-attention. This layer is similar to encoder-decoder attention in standard transformer architecture, except it attends to the output of gated fusion instead of the output of the text encoder.




We observe that the model was able to achieve only up to 0.25 BLEU improvements in some subsets compared to the text-only model, and it decreased for other test sets by up to 1.38 BLEU. The results indicate that the images are not very useful for generating better translation in high-resource settings. Therefore, we experiment with another architecture for Multimodal Translation.

\begin{figure*}
    \centering
    \includegraphics[width=\textwidth]{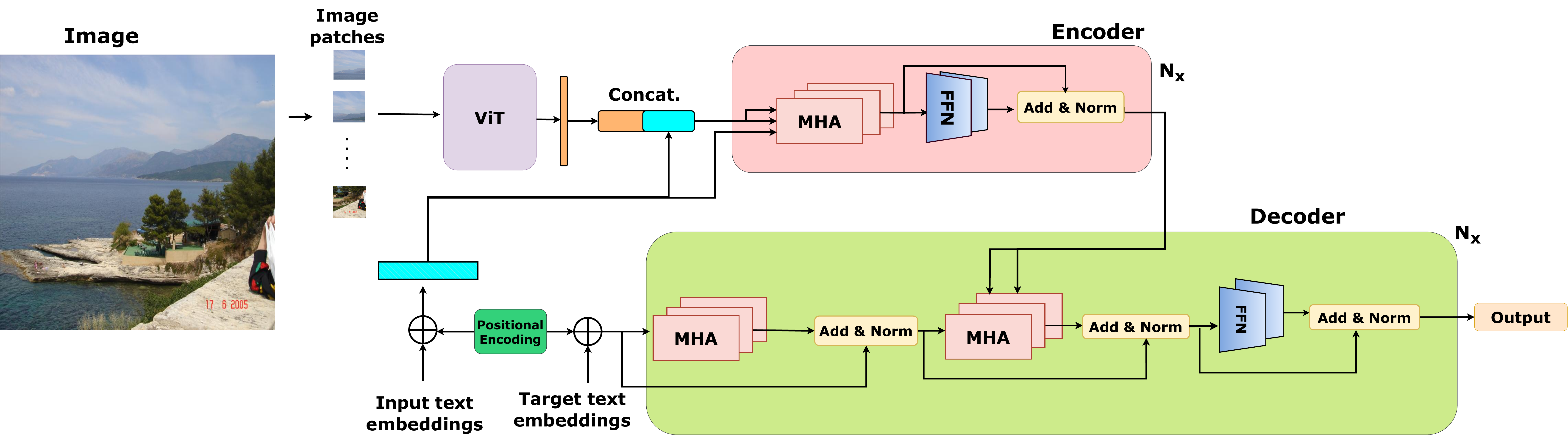}
    \caption{Multimodal Transformer Architecture for Multimodal MT. (softcopy after zooming-in exhibits better display)}
    \label{fig:multimodal-transformer}
\end{figure*}

\subsection{Multimodal Transformer}
\label{subsec:multimodal-transformer}
\begin{table*}[]
\caption{Results of the methods with large-scale pre-training}
\label{tab:result-non-noisy}

\resizebox{0.8\textwidth}{!}{\begin{tabular}{c|c|cc|cc|cc|cc}
\hline
{\multirow{2}{*}{\textbf{Method}}} & \multirow{2}{*}{\textbf{Image Type}} & \multicolumn{2}{c|}{\textbf{Hindi}} & \multicolumn{2}{c|}{\textbf{Bengali}} & \multicolumn{2}{c|}{\textbf{Malayalam}} & \multicolumn{2}{c}{\textbf{Average}} \\
\cline{3-10}
 & & \multicolumn{1}{c}{\textbf{Test}} & \multicolumn{1}{c|}{\textbf{Chal}} & \multicolumn{1}{c}{\textbf{Test}} & \multicolumn{1}{c}{\textbf{Chal}} & \multicolumn{1}{c}{\textbf{Test}} & \multicolumn{1}{c}{\textbf{Chal}} & \multicolumn{1}{c}{\textbf{Test}} & \multicolumn{1}{c}{\textbf{Chal}} \\ 
 \hline \hline
SOTA \citep{parida-etal-2022-silo,gupta-etal-2021-vita}  & Object Tags & 44.64 & 51.66 & 43.9 & 32.9 & 41.00 & 20.4 & 43.18 & 34.99 \\
IndicTrans & None & 34.74 & 46.32 & 26.68 & 28.91 & 28.62 & 26.30 & 30.01 & 33.84 \\
+ Text & None & 45.79 & \textbf{56.72} & 50.08 & 47.78 & 51.38 & \textbf{40.76} & 49.08 & \textbf{48.42} \\
+ SelAttn & Crop & \textbf{46.04} & 56.36 & 48.70 & \textbf{47.82} & 51.56 & 39.76 & 48.77 & 47.98 \\
+ SelAttn & Full & 45.30 & 56.03 & 49.74 & 47.48 & 51.55 & 39.49 & 48.86 & 47.67 \\
+ MMtrans & Crop & 45.47 & 55.61 & 49.03 & 47.32 & \textbf{52.40} & 40.40 & 48.97 & 47.78 \\
+ MMtrans & Full & 45.10 & 56.59 & \textbf{50.45} & 47.24 & 52.00 & 39.77 & \textbf{49.18} & 47.87 \\ \hline
\multicolumn{10}{r}{\scriptsize 
\textbf{Test}: Evaluation Subset; 
\textbf{Chal}: Challenge Subset; 
\textbf{Text}: Text-only finetune; 
SelAttn: Selective Attention; 
MMtrans: Multimodal Transformer}
\end{tabular}}
\end{table*}
We experiment with the Multimodal Transformer \citep{yao-wan-2020-multimodal} as the alternate architecture for the same task. The source text features $x^{text}\in \mathbb{R}^{m\times d}$ are obtained by enriching the source text embedding by the positional embeddings at first. The corresponding image’s features $x^{img}$ are extracted by passing the input image through the Vision Transformer. Since the image and text features belong to different embedding spaces, we project visual features into the text embedding dimension by a learnable projection matrix, $W^{img}$ to obtain $x^{img}W^{img} \in{\mathbb{R}^{n\times d}}$.
Following this, the features are concatenated with each other to obtain a final input representation $x_{com}$ in $R^{(m+n)\times d}$, where projected visual features are $x^{img}W^{img} \in{\mathbb{R}^{n\times d}}$.

In the succeeding phases, we use the combined feature $x_{com}$ as the <key and value> vector for the attention computation in the transformer encoder. The text representation is used as the query vector in the text encoder. 
The output of the final encoder from the last encoder layer is forwarded as keys and values, while the target text representation is passed as the query vector to the MHA module in the decoder module. The encoder and decoder architectures are inspired by the traditional transformer architecture itself. The detail of the architecture is described in \autoref{fig:multimodal-transformer}.
\begin{equation}
    	attn_{ij} = \sum_{j=1}^{m} {softmax}\left(\frac{x^{com}_{i}Q{(x^{text}_{j}K)^{\textrm{T}}}}{\sqrt{d_k}}\right)x^{text}_{j}V; 
\end{equation}
where, $attn_{ij}$ is the output attention between $x^{com}_{i}$ and $x^{text}_{j}$. 
Here, $Q$, $K$, and $V$ are the layer-specific query, key, and value vectors.
We obtain the final encoder representation at the last encoder layer and forward the output to the decoder.

Analogous to the results obtained using the Selective Attention architecture, the Multimodal Transformer architecture was able to achieve only 0.10 BLEU improvements on the test set, on average. However, for the case of the Challenge set, the results actually deteriorated, with a drop of around 0.7 BLEU score. 
\section{Experiments}
\label{sec:results}

This section discusses the results with non-noisy, low-noisy, and high-noisy settings. Further, we report the findings after we replace the actual image with a random image from the dataset; we probe the gate values indicating the relative contribution of visual features compared to textual features. We also report the results by using CLIP as an alternative model for visual feature extraction.

\subsection{Non-Noisy Settings}

\begin{table*}[]
\caption{Difference in results obtained by Multimodal models and Text-based model under non-noisy setting}
\label{tab:diff_clean_multimodal_text}
\begin{tabular}{l|l|rr|rr|rr|rr}
\hline
 &  & \multicolumn{2}{c|}{\textbf{Hindi}} & \multicolumn{2}{c|}{\textbf{Bengali}} & \multicolumn{2}{c|}{\textbf{Malayalam}}   & \multicolumn{2}{c}{\textbf{Average}} \\ \cline{3-10} 
{\multirow{-2}{*}{\textbf{Architecture}}} & \multicolumn{1}{c|}{\multirow{-2}{*}{\textbf{Image Feat}}} & \multicolumn{1}{c|}{\textbf{Eval}} & \multicolumn{1}{c}{\textbf{Chal}} & \multicolumn{1}{c|}{\textbf{Eval}} & \multicolumn{1}{c}{\textbf{Chal}} & \multicolumn{1}{c|}{\textbf{Eval}}   & \multicolumn{1}{c}{\textbf{Chal}} & \multicolumn{1}{c|}{\textbf{Eval}} & \multicolumn{1}{c}{\textbf{Chal}} \\ 
\hline \hline
SelAttn   & Crop   & \multicolumn{1}{r|}{\cellcolor[HTML]{D6DB96}0.25}  & \cellcolor[HTML]{F8C98F}-0.36   & \multicolumn{1}{r|}{\cellcolor[HTML]{E67C73}-1.38} & \cellcolor[HTML]{F9E499}0.04    & \multicolumn{1}{r|}{\cellcolor[HTML]{E2DE97}0.18} & \cellcolor[HTML]{EC987D}-1.00   & \multicolumn{1}{r|}{\cellcolor[HTML]{F9CD90}-0.31} & \cellcolor[HTML]{F7C38C}-0.44   \\ \hline
SelAttn   & Full   & \multicolumn{1}{r|}{\cellcolor[HTML]{F6BF8B}-0.49} & \cellcolor[HTML]{F2B086}-0.69   & \multicolumn{1}{r|}{\cellcolor[HTML]{F8CB8F}-0.34} & \cellcolor[HTML]{F9CE90}-0.30   & \multicolumn{1}{r|}{\cellcolor[HTML]{E4DF97}0.17} & \cellcolor[HTML]{E78476}-1.27   & \multicolumn{1}{r|}{\cellcolor[HTML]{FBD492}-0.22} & \cellcolor[HTML]{F1AB84}-0.75   \\ \hline
MMtrans   & Crop   & \multicolumn{1}{r|}{\cellcolor[HTML]{F9CC90}-0.32} & \cellcolor[HTML]{EA907A}-1.11   & \multicolumn{1}{r|}{\cellcolor[HTML]{EB957C}-1.05} & \cellcolor[HTML]{F6C18C}-0.46   & \multicolumn{1}{r|}{\cellcolor[HTML]{57BB8A}1.02} & \cellcolor[HTML]{F8C98F}-0.36   & \multicolumn{1}{r|}{\cellcolor[HTML]{FDDC95}-0.11} & \cellcolor[HTML]{F3B487}-0.64   \\ \hline
MMtrans   & Full   & \multicolumn{1}{r|}{\cellcolor[HTML]{F2B086}-0.69} & \cellcolor[HTML]{FCDB95}-0.13   & \multicolumn{1}{r|}{\cellcolor[HTML]{C3D694}0.37}  & \cellcolor[HTML]{F5BB8A}-0.54   & \multicolumn{1}{r|}{\cellcolor[HTML]{99CC90}0.62} & \cellcolor[HTML]{ED997D}-0.99   & \multicolumn{1}{r|}{\cellcolor[HTML]{EFE198}0.10}  & \cellcolor[HTML]{F5BB89}-0.55   \\ \hline
\multicolumn{10}{r}{\scriptsize Green/Positive indicates that the MMT model achieved better results than the UNMT model}
\end{tabular}
\end{table*}

We report our results in \autoref{tab:result-non-noisy} under non-noisy setups. In this setting, both training and testing are done with the original dataset. Text-only baseline achieved 49.08 and 48.42 BLEU scores. While these scores are superior to all the existing models, they are due to large-scale pre-training. Selective attention architecture performs inferior to text-only baseline in non-noisy settings. Although it achieves better results in some subsets, the improvements are insignificant. Subsequently, we report the comparison of the multimodal model with respect to the unimodal model in \autoref{tab:diff_clean_multimodal_text}. We observe that the image has an overall negative effect on the results. Surprisingly, it performs worse on the challenge subset, which was constructed with ambiguous words. However, the presence of an ambiguous word does not necessarily make a sentence ambiguous, as the proper meaning of the word can be inferred from the surrounding words within the sentence.

The results in \autoref{tab:diff_clean_multimodal_text} indicate that the images are mostly redundant when used on top of a strong baseline under non-noisy settings. Our observation is similar to \cite{gupta-etal-2021-vita}, where object tags were concatenated with source tags and trained on top of the mBART model. They achieved 0.52 BLEU improvement on the test set over the text-only model. However, BLEU decreased slightly by 0.06 points on the challenge subset for English-Hindi. Although another study \cite{parida-etal-2022-silo} used similar methods and observed huge improvements over text-based models, their baseline was much weaker. Our text-based baseline (without fine-tuning on multimodal data) performs better than their fine-tuned model in some subsets in terms of BLEU score.
We found that images do not have a clear positive impact on the results under non-noisy settings. We investigate if the visual context has a positive impact when part of the source text is deprived or distorted via the addition of noise. Specifically, we add noise types that are predominant in social media text such that multimodal models could be useful for text translation in social media.

\subsection{Low Noise Settings}
The results for low noise settings are reported in \autoref{tab:result-low-noise}. Selective attention with cropped image features surpassed the text-based models on three subsets (out of six). It is to be noted that the improvements were observed on the test subset, where words were not so ambiguous. It generated lower scores in the challenge subset than the text-based model, which contains ambiguous words. This indicates that text-based models are very good at handling most ambiguity when trained on large data.

\begin{table*}[]
\caption{Results of the methods in low-noise settings}
\label{tab:result-low-noise}
\resizebox{0.8\textwidth}{!}{\begin{tabular}{c|c|cc|cc|cc|cc}
\hline
\multicolumn{1}{c|}{\multirow{2}{*}{\textbf{Method}}} & \multirow{2}{*}{\textbf{Image Type}} & \multicolumn{2}{c|}{\textbf{Hindi}} & \multicolumn{2}{c|}{\textbf{Bengali}} & \multicolumn{2}{c|}{\textbf{Malayalam}} & \multicolumn{2}{c}{\textbf{Average}} \\
\cline{3-10}
\multicolumn{1}{c|}{} & & \multicolumn{1}{c}{\textbf{Test}} & \multicolumn{1}{c|}{\textbf{Chal}} & \multicolumn{1}{c}{\textbf{Test}} & \multicolumn{1}{c|}{\textbf{Chal}} & \multicolumn{1}{c}{\textbf{Test}} & \multicolumn{1}{c|}{\textbf{Chal}} & \multicolumn{1}{c}{\textbf{Test}} & \multicolumn{1}{c}{\textbf{Chal}} \\ 
\hline \hline
Text & None & 44.03 & 50.82 & 45.29 & \textbf{42.67} & 47.70 & \textbf{34.76} & 45.67 & \textbf{42.75} \\
SelAttn & Crop & \textbf{44.20} & 50.62 & \textbf{46.07} & 42.33 & \textbf{49.34} & 34.19 & \textbf{46.54} & 42.38 \\
SelAttn & Full & 43.48 & \textbf{51.03} & 45.24 & 41.72 & 47.83 & 34.32 & 45.52 & 42.36 \\
MMtrans & Crop & 43.25 & 50.42 & 45.11 & 42.20 & 48.48 & 34.08 & 45.61 & 42.23 \\
MMtrans & Full & 42.80 & 50.68 & 44.87 & 41.12 & 47.91 & 33.82 & 45.19 & 41.87 \\ \cline{1-10}

\multicolumn{10}{r}{\scriptsize \textbf{Test}: Evaluation subset; \textbf{Chal}: Challenge subset; Text: Text-only fine-tune; SelAttn: Selective Attention; MMtrans: Multimodal Transformer} 
\end{tabular}}
\end{table*}

\begin{table*}[]
\caption{Difference in results obtained by multimodal models and UNMT model under noisy settings}
\label{tab:diff_noise_multimodal_text}
\begin{tabular}{l|l|l|rr|rr|rr|rr}
\hline
\multicolumn{1}{c|}{} & \multicolumn{1}{c|}{} & \multicolumn{1}{c|}{}    & \multicolumn{2}{c|}{\textbf{Hindi}} & \multicolumn{2}{c|}{\textbf{Bengali}}  & \multicolumn{2}{c|}{\textbf{Malayalam}}  & \multicolumn{2}{c}{\textbf{Average}}  \\ \cline{4-11} 
\multicolumn{1}{c|}{\multirow{-2}{*}{\textbf{Architecture}}} & \multicolumn{1}{c|}{\multirow{-2}{*}{\textbf{Image Feat}}} & \multicolumn{1}{c|}{\multirow{-2}{*}{\textbf{Noise}}} & \multicolumn{1}{c|}{\textbf{Eval}} & \multicolumn{1}{c|}{\textbf{Chal}} & \multicolumn{1}{c|}{\textbf{Eval}} & \multicolumn{1}{c|}{\textbf{Chal}} & \multicolumn{1}{c|}{\textbf{Eval}} & \multicolumn{1}{c|}{\textbf{Chal}} & \multicolumn{1}{c|}{\textbf{Eval}} & \multicolumn{1}{c}{\textbf{Chal}} \\ 
\hline \hline
SelAttn   & Crop   & Low & \multicolumn{1}{r|}{\cellcolor[HTML]{EEEDC6}0.17}  & \cellcolor[HTML]{FBE2C0}-0.20   & \multicolumn{1}{r|}{\cellcolor[HTML]{B0D8AD}0.78}  & \cellcolor[HTML]{F9D8B8}-0.34   & \multicolumn{1}{r|}{\cellcolor[HTML]{57BB8A}1.64}  & \cellcolor[HTML]{F5C6AB}-0.57   & \multicolumn{1}{r|}{\cellcolor[HTML]{A6D5A9}0.87}  & \cellcolor[HTML]{F9D5B6}-0.37   \\ \hline
SelAttn   & Full   & Low & \multicolumn{1}{r|}{\cellcolor[HTML]{F6C8AC}-0.55} & \cellcolor[HTML]{EAEBC4}0.21    & \multicolumn{1}{r|}{\cellcolor[HTML]{FEEEC9}-0.05} & \cellcolor[HTML]{EFA995}-0.95   & \multicolumn{1}{r|}{\cellcolor[HTML]{F2EEC7}0.13}  & \cellcolor[HTML]{F7D0B2}-0.44   & \multicolumn{1}{r|}{\cellcolor[HTML]{FCE6C3}-0.15} & \cellcolor[HTML]{F8D4B5}-0.39   \\ \hline
MMtrans   & Crop   & Low & \multicolumn{1}{r|}{\cellcolor[HTML]{F2B69F}-0.78} & \cellcolor[HTML]{F8D3B5}-0.40   & \multicolumn{1}{r|}{\cellcolor[HTML]{FCE4C1}-0.18} & \cellcolor[HTML]{F7CEB1}-0.47   & \multicolumn{1}{r|}{\cellcolor[HTML]{B0D8AD}0.78}  & \cellcolor[HTML]{F4BEA4}-0.68   & \multicolumn{1}{r|}{\cellcolor[HTML]{FEEDC8}-0.06} & \cellcolor[HTML]{F6CAAE}-0.52   \\ \hline
MMtrans   & Full   & Low & \multicolumn{1}{r|}{\cellcolor[HTML]{EB9485}-1.23} & \cellcolor[HTML]{FCE7C3}-0.14   & \multicolumn{1}{r|}{\cellcolor[HTML]{F8D2B3}-0.42} & \cellcolor[HTML]{E67C73}-1.55   & \multicolumn{1}{r|}{\cellcolor[HTML]{EAEBC4}0.21}  & \cellcolor[HTML]{EFAA96}-0.94   & \multicolumn{1}{r|}{\cellcolor[HTML]{F7CDB0}-0.48} & \cellcolor[HTML]{F0AF99}-0.88   \\ \hline
SelAttn   & Crop   & High & \multicolumn{1}{l|}{\cellcolor[HTML]{F8F0CA}0.07}  & \multicolumn{1}{l|}{\cellcolor[HTML]{F6CBAE}-0.51} & \multicolumn{1}{l|}{\cellcolor[HTML]{FDE9C5}-0.11} & \multicolumn{1}{l|}{\cellcolor[HTML]{F2B69F}-0.78} & \multicolumn{1}{l|}{\cellcolor[HTML]{FADDBC}-0.27} & \multicolumn{1}{l|}{\cellcolor[HTML]{EB9787}-1.19} & \multicolumn{1}{l|}{\cellcolor[HTML]{FDEAC6}-0.1}  & \multicolumn{1}{l|}{\cellcolor[HTML]{F1B39C}-0.82} \\ \hline
SelAttn   & Full   & High & \multicolumn{1}{l|}{\cellcolor[HTML]{FEEEC9}-0.04} & \multicolumn{1}{l|}{\cellcolor[HTML]{F6CAAE}-0.52} & \multicolumn{1}{l|}{\cellcolor[HTML]{D2E4BB}0.44}  & \multicolumn{1}{l|}{\cellcolor[HTML]{FCE5C2}-0.17} & \multicolumn{1}{l|}{\cellcolor[HTML]{D2E4BB}0.44}  & \multicolumn{1}{l|}{\cellcolor[HTML]{FCE7C3}-0.14} & \multicolumn{1}{l|}{\cellcolor[HTML]{E3E9C1}0.28}  & \multicolumn{1}{l|}{\cellcolor[HTML]{FADDBC}-0.27} \\ \hline
\multicolumn{11}{r}{\scriptsize Green/Positive indicates that the MMT model achieved better results than the UNMT model}
\end{tabular}
\end{table*}

\begin{table*}[]
\caption{Results of the methods in high-noise settings}
\label{tab:result-high-noise}
\resizebox{0.8\textwidth}{!}{\begin{tabular}{c|c|cc|cc|cc|cc}
\hline
\multicolumn{1}{c|}{\multirow{2}{*}{\textbf{Method}}} & \multirow{2}{*}{\textbf{Image Type}} & \multicolumn{2}{c|}{\textbf{Hindi}} & \multicolumn{2}{c|}{\textbf{Bengali}} & \multicolumn{2}{c|}{\textbf{Malayalam}} & \multicolumn{2}{c}{\textbf{Average}} \\
\cline{3-10}
\multicolumn{1}{c|}{} & & \multicolumn{1}{c}{\textbf{Test}} & \multicolumn{1}{c|}{\textbf{Chal}} & \multicolumn{1}{c}{\textbf{Test}} & \multicolumn{1}{c|}{\textbf{Chal}} & \multicolumn{1}{c}{\textbf{Test}} & \multicolumn{1}{c|}{\textbf{Chal}} & \multicolumn{1}{c}{\textbf{Test}} & \multicolumn{1}{c}{\textbf{Chal}} \\ 
\hline \hline
Text & None & 41.44 & \textbf{45.28} & 42.49 & \textbf{36.56} & 44.38 & \textbf{29.35} & 42.77 & \textbf{37.06} \\
SelAttn & Crop & \textbf{41.51} & 44.77 & 42.38 & 35.78 & 44.11 & 28.16 & 42.67 & 36.24 \\
SelAttn & Full & 41.40 & 44.76 & \textbf{42.93} & 36.39 & \textbf{44.82} & 29.21 & \textbf{43.05} & 36.79 \\ \cline{1-10}
\multicolumn{10}{r}{\scriptsize \textbf{Test}: Evaluation subset; \textbf{Chal}: Challenge subset; Text: Text-only fine-tune; SelAttn: Selective Attention} 
\end{tabular}}
\end{table*}
We report the difference between the multimodal and unimodal models on rows 1-4 of \autoref{tab:diff_noise_multimodal_text}. The selective attention model with cropped image feature outperformed the text-only model by 0.87 BLEU on average on the test set. While multimodal transformer architecture performed competitively with the text-only baseline in some subsets, the overall results were lower than that of the selective attention model on both the challenge and test sets as depicted in \autoref{tab:result-low-noise}. We observe that using only cropped image features has only a positive effect on the evaluation set (row 1) on low-noise settings. Similarly, the cropped image features have a higher positive impact on the Malayalam test set as well as a lesser overall negative impact than full images in multimodal transformer architecture. This suggests the superiority of cropped image features under low-noise settings.

\subsection{High Noise Settings}
Since the Multimodal Transformer model was unable to score the best results in any of the subsets in low-noise settings, we primarily focused on Selective Attention and text-only models.
We report the results in \autoref{tab:result-high-noise} and the difference with respect to the unimodal model on rows 5-6 of \autoref{tab:diff_noise_multimodal_text}. Here, using the cropped image was not found to be very effective compared to the text-based baseline. Selective Attention architecture with cropped images was able to surpass the text-only baseline on only one test set out of six, where the BLEU improvement was only 0.07 points. Using visual features from the full image was more effective than using cropped image features. This is the opposite for low-noise settings. This indicates that the model utilizes visual features more in high-noise settings.

\subsection{Probing with Random Images}

\begin{table*}[]
\caption{Comparison of using Random Images vs. UNMT under noisy setup}
\label{tab:comparsion-noisy-random-unimodal}
\begin{tabular}{l|l|l|cc|cc|cc|cc}
\hline
 & \multicolumn{1}{c|}{} & \multicolumn{1}{c|}{}    & \multicolumn{2}{c|}{\textbf{Hindi}}   & \multicolumn{2}{c|}{\textbf{Bengali}}    & \multicolumn{2}{c|}{\textbf{Malayalam}}  & \multicolumn{2}{c}{\textbf{Average}}    \\ \cline{4-11} 
{\multirow{-2}{*}{\textbf{Architecture}}} & \multicolumn{1}{c|}{\multirow{-2}{*}{\textbf{Image Feat}}} & \multicolumn{1}{c|}{\multirow{-2}{*}{\textbf{Noise}}} & \multicolumn{1}{c|}{\textbf{Eval}} & \textbf{Chal} & \multicolumn{1}{c|}{\textbf{Eval}} & \textbf{Chal} & \multicolumn{1}{c|}{\textbf{Eval}} & \textbf{Chal} & \multicolumn{1}{c|}{\textbf{Eval}} & \textbf{Chal} \\ 
\hline \hline
SelAttn   & Crop   & Low & \multicolumn{1}{c|}{\cellcolor[HTML]{DDDD96}0.26}  & \cellcolor[HTML]{F6C18C}-0.70 & \multicolumn{1}{c|}{\cellcolor[HTML]{6AC08C}1.13}  & \cellcolor[HTML]{F5BC8A}-0.79 & \multicolumn{1}{c|}{\cellcolor[HTML]{57BB8A}1.27}  & \cellcolor[HTML]{F5BB89}-0.81 & \multicolumn{1}{c|}{\cellcolor[HTML]{8AC88F}0.89}  & \cellcolor[HTML]{F5BD8A}-0.77 \\ \hline
SelAttn   & Full   & Low & \multicolumn{1}{c|}{\cellcolor[HTML]{FAD091}-0.40} & \cellcolor[HTML]{FAD292}-0.36 & \multicolumn{1}{c|}{\cellcolor[HTML]{FCDC95}-0.17} & \cellcolor[HTML]{F0A682}-1.22 & \multicolumn{1}{c|}{\cellcolor[HTML]{F3B587}-0.93} & \cellcolor[HTML]{E67C73}-2.05 & \multicolumn{1}{c|}{\cellcolor[HTML]{F8CB8F}-0.50} & \cellcolor[HTML]{F0A782}-1.21 \\ \hline
MMtrans   & Crop   & Low & \multicolumn{1}{c|}{\cellcolor[HTML]{F6C18C}-0.69} & \cellcolor[HTML]{FBD493}-0.32 & \multicolumn{1}{c|}{\cellcolor[HTML]{FCDB95}-0.18} & \cellcolor[HTML]{F9CB8F}-0.49 & \multicolumn{1}{c|}{\cellcolor[HTML]{C0D694}0.48}  & \cellcolor[HTML]{FEE097}-0.08 & \multicolumn{1}{c|}{\cellcolor[HTML]{FDDE96}-0.13} & \cellcolor[HTML]{FBD593}-0.30 \\ \hline
MMtrans   & Full   & Low & \multicolumn{1}{c|}{\cellcolor[HTML]{F5BC8A}-0.79} & \cellcolor[HTML]{F0E298}0.12  & \multicolumn{1}{c|}{\cellcolor[HTML]{F6C08B}-0.72} & \cellcolor[HTML]{ED9C7E}-1.41 & \multicolumn{1}{c|}{\cellcolor[HTML]{7AC48E}1.01}  & \cellcolor[HTML]{F7C68D}-0.60 & \multicolumn{1}{c|}{\cellcolor[HTML]{FDDC96}-0.16} & \cellcolor[HTML]{F7C48D}-0.63 \\ \hline
SelAttn   & Crop   & High & \multicolumn{1}{c|}{\cellcolor[HTML]{FCD894}-0.24} & \cellcolor[HTML]{FBD894}-0.25 & \multicolumn{1}{c|}{\cellcolor[HTML]{FBD493}-0.32} & \cellcolor[HTML]{F4BA89}-0.82 & \multicolumn{1}{c|}{\cellcolor[HTML]{F8CA8F}-0.51} & \cellcolor[HTML]{EFA281}-1.29 & \multicolumn{1}{c|}{\cellcolor[HTML]{FAD292}-0.36} & \cellcolor[HTML]{F5BD8A}-0.78 \\ \hline
SelAttn   & Full   & High & \multicolumn{1}{c|}{\cellcolor[HTML]{EEE198}0.13}  & \cellcolor[HTML]{F9CF91}-0.41 & \multicolumn{1}{c|}{\cellcolor[HTML]{E0DE97}0.24}  & \cellcolor[HTML]{F9CB8F}-0.49 & \multicolumn{1}{c|}{\cellcolor[HTML]{BBD493}0.52}  & \cellcolor[HTML]{F5BB89}-0.81 & \multicolumn{1}{c|}{\cellcolor[HTML]{D8DC96}0.30}  & \cellcolor[HTML]{F8C78E}-0.57 \\ \hline
\multicolumn{11}{r}{\scriptsize Green/Positive indicates that random images achieved better results than the unimodal model}
\end{tabular}
\end{table*}

\begin{table*}[]
\caption{Comparison of using Random Images vs. Actual Images under Noisy setup}
\label{tab:comparsion-noisy-random-nonrandom}
\begin{tabular}{l|l|l|cc|cc|cc|cc}
\hline
 &  &  & \multicolumn{2}{c|}{\textbf{Hindi}}   & \multicolumn{2}{c|}{\textbf{Bengali}}    & \multicolumn{2}{c|}{\textbf{Malayalam}}  & \multicolumn{2}{c}{\textbf{Average}}    \\ \cline{4-11} 
\multicolumn{1}{c|}{\multirow{-2}{*}{\textbf{Architecture}}} & \multicolumn{1}{c|}{\multirow{-2}{*}{\textbf{Image Feat}}} & \multicolumn{1}{c|}{\multirow{-2}{*}{\textbf{Noise}}} & \multicolumn{1}{c|}{\textbf{Eval}} & \textbf{Chal} & \multicolumn{1}{c|}{\textbf{Eval}} & \textbf{Chal} & \multicolumn{1}{c|}{\textbf{Eval}} & \textbf{Chal} & \multicolumn{1}{c|}{\textbf{Eval}} & \textbf{Chal} \\ 
\hline \hline
SelAttn   & Crop   & Low & \multicolumn{1}{c|}{\cellcolor[HTML]{FCD994}-0.09} & \cellcolor[HTML]{CBD895}0.50  & \multicolumn{1}{c|}{\cellcolor[HTML]{F4B788}-0.35} & \cellcolor[HTML]{D1DA95}0.45  & \multicolumn{1}{c|}{\cellcolor[HTML]{D9DC96}0.37}  & \cellcolor[HTML]{E6DF97}0.24  & \multicolumn{1}{c|}{\cellcolor[HTML]{FEE298}-0.02} & \cellcolor[HTML]{D6DB96}0.40  \\ \hline
SelAttn   & Full   & Low & \multicolumn{1}{c|}{\cellcolor[HTML]{FAD191}-0.15} & \cellcolor[HTML]{C4D794}0.57  & \multicolumn{1}{c|}{\cellcolor[HTML]{F3E298}0.12}  & \cellcolor[HTML]{E3DE97}0.27  & \multicolumn{1}{c|}{\cellcolor[HTML]{91CA90}1.06}  & \cellcolor[HTML]{57BB8A}1.61  & \multicolumn{1}{c|}{\cellcolor[HTML]{DBDC96}0.35}  & \cellcolor[HTML]{AAD092}0.82  \\ \hline
MMtrans   & Crop   & Low & \multicolumn{1}{c|}{\cellcolor[HTML]{FCD994}-0.09} & \cellcolor[HTML]{FCDA95}-0.08 & \multicolumn{1}{c|}{\cellcolor[HTML]{FFE599}0.00}  & \cellcolor[HTML]{FDE599}0.02  & \multicolumn{1}{c|}{\cellcolor[HTML]{E0DE97}0.30}  & \cellcolor[HTML]{EC967C}-0.60 & \multicolumn{1}{c|}{\cellcolor[HTML]{F8E499}0.07}  & \cellcolor[HTML]{F8C88E}-0.22 \\ \hline
MMtrans   & Full   & Low & \multicolumn{1}{c|}{\cellcolor[HTML]{F1AB84}-0.44} & \cellcolor[HTML]{F6C28C}-0.26 & \multicolumn{1}{c|}{\cellcolor[HTML]{E0DE97}0.30}  & \cellcolor[HTML]{FAD292}-0.14 & \multicolumn{1}{c|}{\cellcolor[HTML]{E67C73}-0.80} & \cellcolor[HTML]{F4B888}-0.34 & \multicolumn{1}{c|}{\cellcolor[HTML]{F5BB89}-0.32} & \cellcolor[HTML]{F7C48D}-0.25 \\ \hline
SelAttn   & Crop   & High & \multicolumn{1}{c|}{\cellcolor[HTML]{DFDD97}0.31}  & \cellcolor[HTML]{F6C28C}-0.26 & \multicolumn{1}{c|}{\cellcolor[HTML]{EAE098}0.21}  & \cellcolor[HTML]{FBE499}0.04  & \multicolumn{1}{c|}{\cellcolor[HTML]{E6DF97}0.24}  & \cellcolor[HTML]{F5E399}0.10  & \multicolumn{1}{c|}{\cellcolor[HTML]{E4DF97}0.26}  & \cellcolor[HTML]{FDDF97}-0.04 \\ \hline
SelAttn   & Full   & High & \multicolumn{1}{c|}{\cellcolor[HTML]{F9CE90}-0.17} & \cellcolor[HTML]{FBD693}-0.11 & \multicolumn{1}{c|}{\cellcolor[HTML]{EBE098}0.20}  & \cellcolor[HTML]{DEDD97}0.32  & \multicolumn{1}{c|}{\cellcolor[HTML]{FCDA95}-0.08} & \cellcolor[HTML]{BAD493}0.67  & \multicolumn{1}{c|}{\cellcolor[HTML]{FEE298}-0.02} & \cellcolor[HTML]{E0DE97}0.30  \\ \hline
\multicolumn{11}{r}{\scriptsize Green/Positive indicates that actual images achieved better results than random images}
\end{tabular}
\end{table*}

At first glance, the comparable performance of models using random and relevant images may seem to challenge the assertion that visual context is beneficial for handling noisy inputs. However, a deeper examination reveals a more nuanced interaction between visual context and model behavior. Our experiments demonstrate that incorporating visual context consistently improves performance in noisy settings compared to text-only models, affirming that visual information is indeed "helpful." However, the observation that random images yield similar results to relevant ones suggests that the observed improvements are not attributable to the semantic relevance of the visual information. Instead, the inclusion of image features—whether random or relevant—appears to enhance model robustness, potentially through regularization or by mitigating the effects of textual noise.

To investigate whether these improvements stem from meaningful visual information, we conducted probing experiments where relevant images were replaced with randomly selected images from the same dataset. The results, as presented in \autoref{tab:comparsion-noisy-random-nonrandom}, reveal that in low-noise settings, random images yield similar performance to relevant images for cropped features. For instance, the Selective Attention architecture achieved comparable results on the test set with random images, with only a marginal drop of -0.40 BLEU on the challenge set. However, when using full images, the use of random images had a more pronounced effect, reducing BLEU by 0.35 and 0.82 on the Test and Challenge subsets, respectively. This outcome aligns with the observation that textual context alone is often sufficient in low-noise scenarios, as evidenced by the strong performance of text-only models fine-tuned on noisy data.

In high-noise settings, the results with random images remained largely consistent with those obtained using relevant images. Comparisons with text-based baselines, as detailed in \autoref{tab:comparsion-noisy-random-unimodal}, exhibit trends similar to those in \autoref{tab:diff_noise_multimodal_text}. These findings indicate that the performance gains achieved with relevant images can also be attained using random images. This suggests that the model does not leverage meaningful visual information but instead utilizes image features as a form of noise, enhancing its robustness during training. Consequently, the absence of correct images during inference has minimal impact on performance.

These results highlight a critical limitation of current multimodal approaches: while visual context improves overall performance, the model appears to rely on generic signal patterns from image features rather than extracting and integrating complementary information. This underscores the need for further research to develop mechanisms that can effectively extract and leverage relevant visual features, ensuring that the visual modality contributes meaningfully rather than merely acting as a source of noise.

\subsection{Using CLIP image features}
\begin{table*}[]
\caption{Results of CLIP features vs Vision Transformer feature in high-noise settings}
\label{tab:result-high-noise-clip-vs}
\resizebox{0.8\textwidth}{!}{\begin{tabular}{c|c|cc|cc|cc|cc}
\hline
{\multirow{2}{*}{\textbf{Method}}} & \multirow{2}{*}{\textbf{Image Feature}} & \multicolumn{2}{c|}{\textbf{Hindi}} & \multicolumn{2}{c|}{\textbf{Bengali}} & \multicolumn{2}{c|}{\textbf{Malayalam}} & \multicolumn{2}{c}{\textbf{Average}} \\
\cline{3-10}
 & & \multicolumn{1}{c}{\textbf{Test}} & \multicolumn{1}{c|}{\textbf{Chal}} & \multicolumn{1}{c}{\textbf{Test}} & \multicolumn{1}{c|}{\textbf{Chal}} & \multicolumn{1}{c}{\textbf{Test}} & \multicolumn{1}{c|}{\textbf{Chal}} & \multicolumn{1}{c}{\textbf{Test}} & \multicolumn{1}{c}{\textbf{Chal}} \\ 
\hline \hline
Text & None & 41.44 & \textbf{45.28} & 42.49 & \textbf{36.56} & 44.38 & \textbf{29.35} & 42.77 & \textbf{37.06} \\
SelAttn & CLIP & 41.11 & 45.21 & \textbf{42.67} & 36.35 & 44.61 & 29.24 & \textbf{42.80} & 36.93 \\
SelAttn & ViT & \textbf{41.51} & 44.77 & 42.38 & 35.78 & 44.11 & 28.16 & 42.67 & 36.24 \\  \cline{1-10}

\multicolumn{10}{r}{\scriptsize \textbf{Test}: Evaluation subset; \textbf{Chal}: Challenge subset; Text: Text-only fine-tune; SelAttn: Selective Attention} 
\end{tabular}}
\end{table*}
We explore whether the relatively low impact of image features in multimodal translation can be attributed to the type of features used in previous experiments. To deepen our understanding, we extend the investigation by incorporating image features derived from CLIP \cite{radford2021learningtransferablevisualmodels}, a state-of-the-art model designed for learning transferable visual representations. We chose to perform the experiment on high-noise settings because high-noise settings have been found to be most effective for utilizing image features. We use ViT-B/16 model to extract image features and use it as $H_{img}$ in \autoref{equation:1}. Our findings, summarized in \autoref{tab:result-high-noise-clip-vs}, reveal that the use of CLIP-derived image features does not significantly alter the overall observations. The model's performance with these features remains closely aligned with that achieved using ViT-based features from earlier experiments. This consistency underscores a notable insight: the limited influence of image features in multimodal translation may not primarily stem from the feature extraction.

\subsection{Gate Values}
Given the observed limited contribution of images to the translation quality, we further investigated the behavior of the gating mechanism ($\lambda$), as described in \autoref{eq:gate}. The gate value, which dynamically adjusts based on the image and text representations, provides insight into the relative importance assigned to the image modality during translation.

In the high-noise training regime with the full image settings, the average gate values computed over the test set and challenge set were found to be 0.000781 and 0.000742, respectively. Interestingly, when random images were used during inference, the corresponding average gate values were 0.000780 and 0.000738 for the test and challenge sets. These values demonstrate minimal variation between the settings, indicating that the model assigns an almost negligible weight to the image modality, regardless of whether the input images are relevant or random. This consistency further supports our hypothesis that the contribution of the image modality is more aligned with regularization rather than providing substantive contextual information to improve translation quality. 
\subsection{Using Multi30K dataset for Testing}
The primary focus of this work is on multimodal translation from English to Indic languages. The widely used Multi30K dataset \cite{elliott-etal-2016-multi30k} has been used in several methods to generate multimodal translation \cite{peng-etal-2022-distill}, visual pre-training \cite{shan2022ernieunix2}, image captioning \cite{9878503}, etc. However, this dataset supports German, French, and English. Moreover, the pre-trained models employed in our experiments are not tailored for the languages in the Multi30K dataset, making direct comparisons challenging. Despite these constraints, we conducted an exploratory analysis as follows.

We utilized the Multi30K test set to generate translations using our trained models under high-noise settings. Since this dataset lacks references in Indic languages, we relied on human annotations to evaluate the quality of the translations.
For each test instance, two hypotheses were generated:
\begin{enumerate}
    \item[h1:] Translations produced using the actual image associated with the text.
    \item[h2:] Translations produced using a randomly sampled image during inference.
\end{enumerate}

\noindent
A subset of 50 English source sentences from the Multi30K dataset was randomly selected, and translations were evaluated in both the English-Hindi and English-Bengali directions. Two experienced annotators were tasked with comparing the translations for each hypothesis and selecting one of the following options:

\begin{itemize}
    \item Hypothesis-h1 is much better (enhances adequacy of the translation).
    \item Hypothesis-h1 is better (with less significant or minor grammatical improvements).
    \item Both hypotheses are of similar quality.
    \item Hypothesis-h2 is much better (enhances adequacy of the translation).
    \item Hypothesis-h2 is better (with less significant or minor grammatical improvements).
\end{itemize}

\begin{table}[h!]
\centering
\caption{Evaluation results comparing translations produced with actual and random images on Multi30k Test set.}
\label{tab:results-multi30k}
\begin{adjustbox}{width=0.8\linewidth}   
\begin{tabular}{l|c|c}
\hline
\textbf{Option} & \textbf{English-Hindi} & \textbf{English-Bengali} \\ \hline \hline
Hypothesis-h1 is much better (enhances adequacy) & 12 & 10 \\ \hline
Hypothesis-h1 is better (minor grammatical improvements) & 4 & 2 \\ \hline
Both hypotheses are of similar quality & 15 & 25 \\ \hline
Hypothesis-h2 is much better (enhances adequacy) & 13 & 11 \\ \hline
Hypothesis-h2 is better (minor grammatical improvements) & 6 & 2 \\ \hline
\end{tabular}
\end{adjustbox}
\end{table}

The results of the human evaluation are summarized in \autoref{tab:results-multi30k}. The findings reveal no significant difference between translations generated using actual images and those generated with randomly sampled images. For English-Hindi, translations with actual images are rated better in 12 instances, whereas translations with random images are preferred in 13 instances. Similar trends are observed for English-Bengali. These results support our hypothesis that images primarily act as regularizers rather than providing significant contextual information for improving translation quality.

It is important to note that the overall quality of translations in this analysis was reported to be poor, likely due to the high-noise settings used during training and inference. This limitation restricted the scope of our evaluation and highlighted the need for further investigation to isolate the impact of images. However, such an exploration is beyond the scope of this paper.

\section{Ablation and Discussion}
\label{sec:ablation}
\begin{table*}[]
\centering
\caption{\textcolor{black}{The results by removing the Selective Attention module entirely (rows 6 to 8) and replacing the combined image and text feature with text features only, removing the image modality (rows 4 to 5)}}
\label{tab:ablation}
\resizebox{0.8\textwidth}{!}{
\begin{tabular}{>{\color{black}}l|>{\color{black}}l|>{\color{black}}l|>{\color{black}}l|>{\color{black}}l|>{\color{black}}l|>{\color{black}}l|>{\color{black}}l|>{\color{black}}l|>{\color{black}}l|>{\color{black}}l}
\hline
\multirow{2}{*}{\textbf{Method}}          & \textbf{Training} & \textbf{Testing} & \multicolumn{2}{c|}{\textbf{\textcolor{black}{Hindi}}}         & \multicolumn{2}{c|}{\textbf{\textcolor{black}{Bengali}}}       & \multicolumn{2}{c|}{\textbf{\textcolor{black}{Malayalam}}}     & \multicolumn{2}{c}{\textbf{\textcolor{black}{Average}}}       \\ \cline{4-11} 
 & \textbf{Features} & \textbf{Features} & \textbf{Test}  & \textbf{Chal}  & \textbf{Test}  & \textbf{Chal}  & \textbf{Test}  & \textbf{Chal}  & \textbf{Test}  & \textbf{Chal}  \\ 
\hline \hline
SelAttn                          & Actual                               & Actual                              & 41.40 & 44.76 & 42.93 & 36.39 & 44.82 & 29.21 & 43.05 & 36.79 \\ 
SelAttn                          & Actual                               & Random                              & 41.53 & 44.35 & 43.17 & 35.90 & 45.34 & 28.40 & 43.35 & 36.22 \\ 
SelAttn                          & Random                               & Random                              & 41.80 & 45.60 & 42.88 & 36.53 & 44.86 & 29.45 & 43.18 & 37.19 \\ 
\hline 
SelAttn $\rightarrow$ Text & Actual                               & None                                & 41.35 & 44.75 & 42.84 & 36.19 & 45.02 & 28.59 & 43.07 & 36.51 \\ 
SelAttn $\rightarrow$ Text & Random                               & None                                & 41.79 & 45.60 & 42.88 & 36.56 & 44.73 & 29.45 & 43.13 & 37.20 \\ 
\hline 
NoSelAttn           & Actual                               & Actual                              & 41.42 & 45.31 & 42.61 & 35.46 & 44.32 & 28.22 & 42.78 & 36.33 \\ 
NoSelAttn           & Actual                               & Random                              & 41.20 & 45.26 & 42.71 & 35.78 & 43.93 & 28.03 & 42.61 & 36.36 \\ 
NoSelAttn           & Random                               & Random                              & 42.08 & 45.36 & 43.70 & 36.57 & 44.67 & 29.08 & 43.48 & 37.00 \\ \hline
\multicolumn{11}{r}{\scriptsize \textbf{Test}: Evaluation subset; \textbf{Chal}: Challenge subset; Text: Text-only fine-tune; SelAttn: Selective Attention; NoSelAttn: Not Selective Attention}
\end{tabular}}
\end{table*}

\textcolor{black}{To assess the contribution of different components in our multimodal architecture, we conduct a detailed ablation study, the results of which are presented in Table~\ref{tab:ablation}.}

\subsection{Impact of Removing Visual Features.}
\textcolor{black}{Rows 4 and 5 in Table \ref{tab:ablation} present results where the multimodal input is reduced to text-only features by removing the visual modality while retaining the rest of the architecture, including the gating mechanism. Specifically, we replace the fused image-text representation with only text-derived features. We experiment with both actual and randomly sampled training features, and in both cases, we remove visual features entirely at test time. Surprisingly, this simplification results in only marginal changes in BLEU scores across all languages and evaluation settings.}

\textcolor{black}{For example, using actual training features and removing image features at test time (row 4), the model achieves an average BLEU score of 43.07 on test sets and 36.51 on challenge sets. This is nearly identical to the performance of the full SelAttn (Selective Attention) mechanism with actual image features (row 1), which scores 43.05 and 36.79, respectively. A similar trend holds when random features are used for training (row 5), with scores of 43.13 (Test) and 37.20 (Challenge). These results suggest that the image modality, in its current integration, contributes little in terms of meaningful semantic enhancement and can be effectively replaced by text-only input without sacrificing performance.}

\subsection{Impact of Removing Selective Attention}
\textcolor{black}{Rows 6\--8 of Table \ref{tab:ablation} evaluate the impact of removing the SelAttn entirely. Here, the gating and fusion modules are retained, but the attention layer that modulates visual information is disabled. Interestingly, this also results in minimal performance degradation. In fact, the best test BLEU average of 43.48 is achieved in the configuration where both training and testing use randomly sampled features without any selective attention (row 8). This slightly outperforms the full SelAttn model, further questioning the efficacy of the attention mechanism in contributing meaningful information from the image stream.}

\textcolor{black}{Moreover, when actual features are used for training and testing without SelAttn (row 6), the performance is 42.78 (Test) and 36.33 (Challenge), again showing only a marginal drop from the baseline. These findings imply that the selective attention mechanism offers limited gains in this setup, and its removal does not adversely affect the system's robustness or accuracy.}

\subsection{Effect of Random Image Features}
\textcolor{black}{Another notable pattern across Table \ref{tab:ablation} is the consistent performance of models that utilize random image features. Whether SelAttn is enabled or not, and regardless of whether training or testing features are randomized, the BLEU scores remain relatively stable. For instance, the SelAttn model with random features at both training and test time (row 3) achieves 43.18 (Test) and 37.19 (Challenge), which are slightly better than the full model with actual features. This observation suggests that the gains observed from using the image stream may not be due to semantic visual information but rather due to the introduction of controlled noise that acts as a regularizer.}

\subsection{Effectiveness of Context for Machine Translation}
\textcolor{black}{Overall, the ablation results indicate that the visual modality and the selective attention mechanism are not strictly necessary for strong performance in our task. Replacing the image features with random noise or removing the attention module altogether does not substantially degrade results. These findings highlight that the benefits attributed to the image modality likely stem from the stochastic regularization effect it introduces rather than from its semantic content. This suggests an alternative use of multimodal inputs: not primarily for semantic grounding, but as a source of structured variability to improve robustness, especially in noisy or low-resource scenarios.}

\textcolor{black}{While the observations are surprising, a growing body of research suggests that the actual contribution of context (here, image) may be overstated. A detailed evaluation designed with ambiguous source sentences reveals that although MMT systems can technically process visual information, they often fail to integrate it meaningfully into translation choices, limiting its practical use for disambiguation or grounding \cite{li-etal-2022-vision,kashani-motlagh-etal-2024-assessing}. 
Adding to this perspective, another study systematically replaced relevant images with random ones in a multimodal prompting setup and observed that translation performance remained largely unaffected \cite{baldassini2024makesmultimodalincontextlearning}. This surprising result indicates that visual features may primarily stabilize learning dynamics rather than provide true contextual relevance, aligning with the hypothesis that random images can offer similar benefits.}

\textcolor{black}{This phenomenon is not unique to multimodal contexts. In document-level and dialogue translation, where external textual context is incorporated, improvements over sentence-level baselines are frequently modest and inconsistent \cite{jin-etal-2023-challenges}. Despite leveraging information from surrounding sentences or dialogue turns, models often fail to utilize this additional context effectively. A broader analysis of context-aware systems identifies a similar trend: while these models occasionally handle specific phenomena better, such as pronoun resolution or discourse coherence, their overall performance remains comparable to simpler sentence-level models \cite{huo-etal-2020-diving}. This raises questions about how much external context truly contributes to translation quality when averaged across benchmarks. Interestingly, several of these studies have hypothesized that external context might be more beneficial when the input text itself is noisy or ambiguous \cite{caglayan-etal-2019-probing}. However, we observe that even in such cases, image features fail to consistently enhance translations. Note that while the previous study used masked tokens to indicate missing information, which may have helped the model to look for additional information, we provide no such hints to the model to simulate real-life noisy examples. 
Collectively, these observations support the emerging view that external context in machine translation often functions more as a form of implicit regularization than as a direct source of disambiguating information. The challenge of effectively integrating this context into translation models remains largely unsolved.}

\begin{figure*}
 \centering
 \fbox{ \includegraphics[width=0.8\textwidth,keepaspectratio]{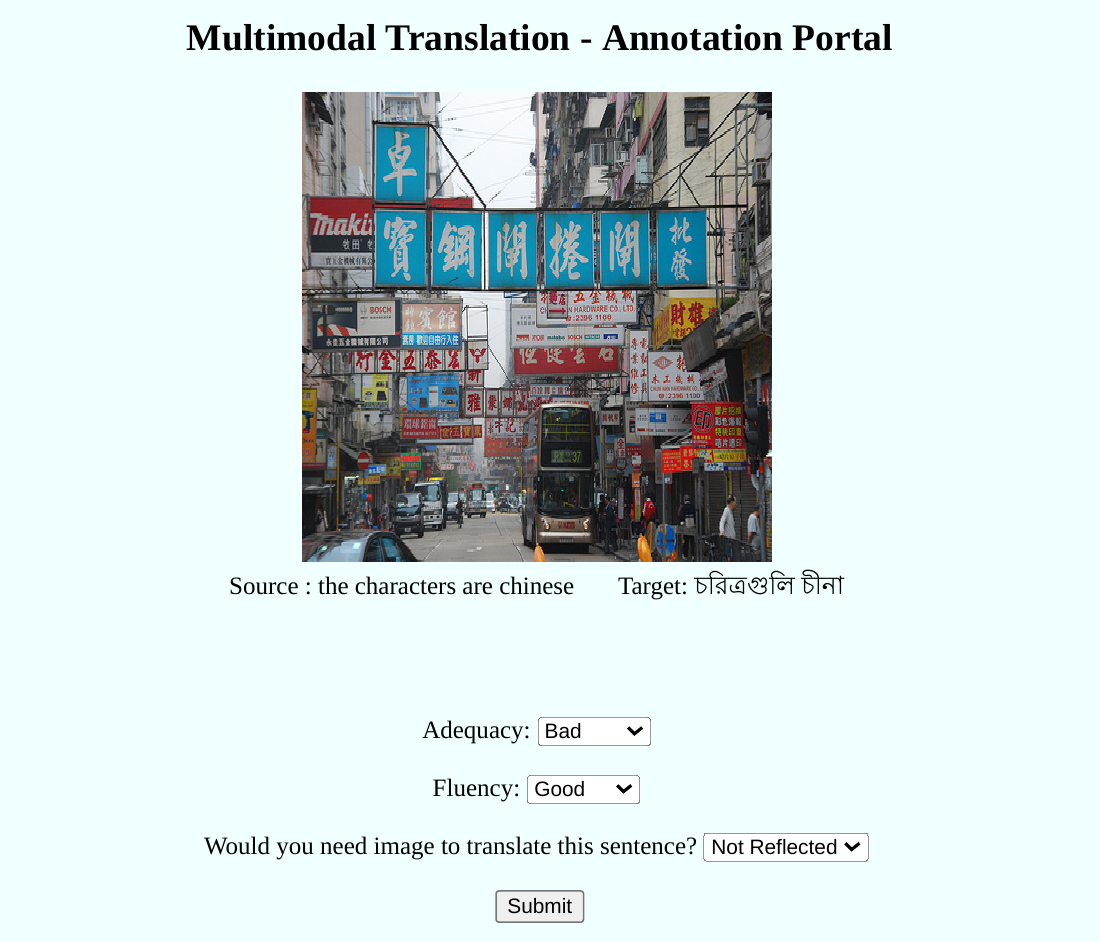}}
 \caption{Example of the annotation process. This example is obtained from the Challenge Subset of Bengali VG. It is to be noted that the reference is wrong since the image indicates that \textit{character} refers to words in the banner. However, the reference is referring the word \textit{character} to \textit{protagonist} (of movies, stories, etc.) }
 \label{fig:annotation_example_bengali}
\end{figure*}

\subsection{Exploring Dataset Quality}

We attempt to find out if the dataset is context-aware enough to facilitate the models to learn from images, and if the test sets are context-aware to truly influence the results if the models are able to generate image-aware translation. We build an annotation web portal to check the adequacy, fluency, and image awareness of the dataset by manual evaluation. We provide the annotators with the source, target, and image. 
We ask the annotator to rate adequacy and fluency on a scale of three - \emph{good, medium, and bad}. 
Further, we ask the annotators if they think the image is needed to generate the translation. The answer should be \emph{No} if they would translate it the same even if the image were absent. The answer should be \emph{Yes} if they would have translated it differently or be in confusion without the availability of the image. The answer should be \emph{maybe} when they would have translated similarly anyway, but the image helped to clear the confusion. Finally, the answer is \emph{not reflected} when they believe that the translation could be better with the information from the image, but the same is not reflected in the reference. 

\begin{table*}[]
\centering
\caption{Details of annotation for quality and image awareness of the datasets}
\label{tab:image-awareness}
\resizebox{0.75\textwidth}{!}{\begin{tabular}{c|c | cc | cc | cc}
\hline
 & & \multicolumn{2}{c|}{\textbf{Challenge Set (\%)}} & \multicolumn{2}{c|}{\textbf{Test Set (\%)}} & \multicolumn{2}{c}{\textbf{Train Set (\%)}} \\
 \cline{3-8} 
 & Good & 82 & 74 & 92 & 88 & 86 & 64 \\
\textbf{Adequacy} & Medium & 8 & 18 & 8 & 8 & 14 & 32 \\
\textbf{} & Bad & 10 & 8 & 0 & 4 & 0 & 4 \\ \cline{1-8}
\textbf{} & Good & 86 & 74 & 80 & 72 & 62 & 48 \\
\textbf{Fluency} & Medium & 14 & 14 & 20 & 22 & 30 & 42 \\
\textbf{} & Bad & 0 & 12 & 0 & 6 & 8 & 10 \\ \cline{1-8}
\textbf{} & Yes & \multicolumn{2}{c|}{6} & \multicolumn{2}{c|}{0} & \multicolumn{2}{c}{0} \\
\textbf{Need of Image} & Maybe & \multicolumn{2}{c|}{4} & \multicolumn{2}{c|}{4} & \multicolumn{2}{c}{10} \\
\textbf{} & No & \multicolumn{2}{c|}{84} & \multicolumn{2}{c|}{94} & \multicolumn{2}{c}{90} \\
\textbf{} & Not Reflected & \multicolumn{2}{c|}{6} & \multicolumn{2}{c|}{2} & \multicolumn{2}{c}{0} \\ \cline{1-8}
\end{tabular}}
\end{table*}

We report our observations in \autoref{tab:image-awareness}. While the dataset is mostly fluent and adequate, the source text is usually enough to translate it. In the train set, 90\% of the examples do not require an image to translate correctly, while in 10\%, it would have been somewhat helpful but can be translated without it. 
This prevents the models from getting proper training signals to use multimodal information effectively.
In the challenge set of VisualGenome (VG) datasets, only 6\% of the dataset would require the image to translate it, whereas 4\% would have been somewhat helpful. This is 0\% and 4\% for the test set.
There are many instances where an image is essential for accurately translating a sentence. However, the reference translation does not adequately capture this image-aware context. We illustrate one such example in \autoref{fig:annotation_example_bengali}. This occurs in 6\% and 2\% of the examples in Challenge and Test sets, respectively.
We generate this report on the first 50 examples of Train, Challenge, and Test Sets of Hindi VG and Bengal VG.
This indicates that only a small percentage of examples require images for correct translation. However, the usage of examples that are difficult to translate would encourage the models to use visual features effectively, even when it is trained on a large unimodal corpus. However, to the best of our knowledge, no such data is publicly available.

\subsection{Experimental Setup}
We try to keep the hyperparameters the same in every model for a fair comparison.
We set the learning rate to $3 \times 10^{-5}$, which performed the best for the baseline model \citep{ramesh-etal-2022-samanantar} during fine-tuning. We set warmup updates to 4000 and the initial learning rate to $10^{-7}$ with inverse square root as the learning rate scheduler. 
Due to the small size of the dataset, we validate and save checkpoints at frequent intervals, at multiples of 500 steps or at the end of an epoch. 
We continue training with Early Stopping till the validation loss does not improve for five consecutive checkpoints. We use \emph{Label Smoothed Cross Entropy} as the loss function and set 0.1 as the \emph{smoothing\_value}. 
We set the \emph{batch\_size} to 256 tokens and \emph{update\_frequency} to 2. 
Effectively, this makes the \emph{batch\_size} of 512 tokens per step. 
We use a single GeForce RTX 2080 Ti with 11 GB GPU RAM. 
All experiments are conducted on Ubuntu 18.04.1 LTS x86\_64. 
We perform memory-efficient FLOP16 training and use the fairseq \citep{ott-etal-2019-fairseq} library for our implementations. 
We set 42 as the \emph{random\_seed} and 0.7 as the \emph{keeping\_probability} of dropout.
During testing, we used beam search with a \emph{beam\_width} of 5.
\section{Conclusion}
\label{conclusion}

Multimodal Neural Machine Translation (NMT) can be useful for translating social media content. Previous studies suggest that multimodal models generate better translations compared to unimodal NMT. However, this does not always hold true, especially when the NMT model is trained on a large corpus. We observe that images are often redundant in such high-resource settings. Our findings show that two well-known Multimodal Machine Translation (MMT) models do not consistently outperform a strong text-based baseline. 
We also find that visual context becomes valuable when the source text is noisy. Experimentally, cropped image features achieve the best results in low-noise settings, while full image features yield the best performance in high-noise settings. This indicates that the impact of visual features increases with the level of noise in the source text. However, we found that the improvements were not due to meaningful information from the images, as similar results were achieved even with random image features. This suggests that current models do not yet use images effectively, even when the text becomes less adequate due to noise. One possible solution is to employ datasets where the image is essential, thereby enabling models to learn to integrate visual features more effectively. 
In future work, we aim to explore the effectiveness of multilingual multimodal NMT across other languages and datasets under noisy conditions.

\section*{Acknowledgments}
Baban Gain and Dibyanayan Bandyopadhyay gratefully acknowledge the Prime Minister’s Research Fellowship (PMRF) program for providing financial support and enabling this research.

\bibliographystyle{ACM-Reference-Format}
\bibliography{samples/sample-base,samples/anthology}


\begin{thebibliography}{86}


\ifx \showCODEN    \undefined \def \showCODEN     #1{\unskip}     \fi
\ifx \showDOI      \undefined \def \showDOI       #1{#1}\fi
\ifx \showISBNx    \undefined \def \showISBNx     #1{\unskip}     \fi
\ifx \showISBNxiii \undefined \def \showISBNxiii  #1{\unskip}     \fi
\ifx \showISSN     \undefined \def \showISSN      #1{\unskip}     \fi
\ifx \showLCCN     \undefined \def \showLCCN      #1{\unskip}     \fi
\ifx \shownote     \undefined \def \shownote      #1{#1}          \fi
\ifx \showarticletitle \undefined \def \showarticletitle #1{#1}   \fi
\ifx \showURL      \undefined \def \showURL       {\relax}        \fi
\providecommand\bibfield[2]{#2}
\providecommand\bibinfo[2]{#2}
\providecommand\natexlab[1]{#1}
\providecommand\showeprint[2][]{arXiv:#2}

\bibitem[Appicharla et~al\mbox{.}(2023)]%
        {appicharla2023case}
\bibfield{author}{\bibinfo{person}{Ramakrishna Appicharla}, \bibinfo{person}{Baban Gain}, \bibinfo{person}{Santanu Pal}, {and} \bibinfo{person}{Asif Ekbal}.} \bibinfo{year}{2023}\natexlab{}.
\newblock \bibinfo{title}{A Case Study on Context Encoding in Multi-Encoder based Document-Level Neural Machine Translation}.
\newblock
\newblock
\showeprint[arxiv]{2308.06063}~[cs.CL]


\bibitem[Baldassini et~al\mbox{.}(2024)]%
        {baldassini2024makesmultimodalincontextlearning}
\bibfield{author}{\bibinfo{person}{Folco~Bertini Baldassini}, \bibinfo{person}{Mustafa Shukor}, \bibinfo{person}{Matthieu Cord}, \bibinfo{person}{Laure Soulier}, {and} \bibinfo{person}{Benjamin Piwowarski}.} \bibinfo{year}{2024}\natexlab{}.
\newblock \bibinfo{title}{What Makes Multimodal In-Context Learning Work?}
\newblock
\newblock
\showeprint[arxiv]{2404.15736}~[cs.CV]
\urldef\tempurl%
\url{https://arxiv.org/abs/2404.15736}
\showURL{%
\tempurl}


\bibitem[Belinkov and Bisk(2018)]%
        {belinkov2018synthetic}
\bibfield{author}{\bibinfo{person}{Yonatan Belinkov} {and} \bibinfo{person}{Yonatan Bisk}.} \bibinfo{year}{2018}\natexlab{}.
\newblock \bibinfo{title}{Synthetic and Natural Noise Both Break Neural Machine Translation}.
\newblock
\newblock
\showeprint[arxiv]{1711.02173}~[cs.CL]


\bibitem[Berard et~al\mbox{.}(2019)]%
        {berard-etal-2019-naver}
\bibfield{author}{\bibinfo{person}{Alexandre Berard}, \bibinfo{person}{Ioan Calapodescu}, {and} \bibinfo{person}{Claude Roux}.} \bibinfo{year}{2019}\natexlab{}.
\newblock \showarticletitle{Naver Labs {E}urope`s Systems for the {WMT}19 Machine Translation Robustness Task}. In \bibinfo{booktitle}{\emph{Proceedings of the Fourth Conference on Machine Translation (Volume 2: Shared Task Papers, Day 1)}}, \bibfield{editor}{\bibinfo{person}{Ond{\v{r}}ej Bojar}, \bibinfo{person}{Rajen Chatterjee}, \bibinfo{person}{Christian Federmann}, \bibinfo{person}{Mark Fishel}, \bibinfo{person}{Yvette Graham}, \bibinfo{person}{Barry Haddow}, \bibinfo{person}{Matthias Huck}, \bibinfo{person}{Antonio~Jimeno Yepes}, \bibinfo{person}{Philipp Koehn}, \bibinfo{person}{Andr{\'e} Martins}, \bibinfo{person}{Christof Monz}, \bibinfo{person}{Matteo Negri}, \bibinfo{person}{Aur{\'e}lie N{\'e}v{\'e}ol}, \bibinfo{person}{Mariana Neves}, \bibinfo{person}{Matt Post}, \bibinfo{person}{Marco Turchi}, {and} \bibinfo{person}{Karin Verspoor}} (Eds.). \bibinfo{publisher}{Association for Computational Linguistics}, \bibinfo{address}{Florence, Italy}, \bibinfo{pages}{526--532}.
\newblock
\urldef\tempurl%
\url{https://doi.org/10.18653/v1/W19-5361}
\showDOI{\tempurl}


\bibitem[Bisht and Solanki(2022)]%
        {10.1007/978-981-19-4831-2_5-caption-translation-object}
\bibfield{author}{\bibinfo{person}{Paritosh Bisht} {and} \bibinfo{person}{Arun Solanki}.} \bibinfo{year}{2022}\natexlab{}.
\newblock \showarticletitle{Exploring Practical Deep Learning Approaches for English-to-Hindi Image Caption Translation Using Transformers and Object Detectors}. In \bibinfo{booktitle}{\emph{Applications of Artificial Intelligence and Machine Learning}}, \bibfield{editor}{\bibinfo{person}{Bhuvan Unhelker}, \bibinfo{person}{Hari~Mohan Pandey}, {and} \bibinfo{person}{Gaurav Raj}} (Eds.). \bibinfo{publisher}{Springer Nature Singapore}, \bibinfo{address}{Singapore}, \bibinfo{pages}{47--60}.
\newblock
\showISBNx{978-981-19-4831-2}


\bibitem[Caglayan et~al\mbox{.}(2016)]%
        {caglayan-etal-2016-multimodality}
\bibfield{author}{\bibinfo{person}{Ozan Caglayan}, \bibinfo{person}{Walid Aransa}, \bibinfo{person}{Yaxing Wang}, \bibinfo{person}{Marc Masana}, \bibinfo{person}{Mercedes Garc{\'i}a-Mart{\'i}nez}, \bibinfo{person}{Fethi Bougares}, \bibinfo{person}{Lo{\"i}c Barrault}, {and} \bibinfo{person}{Joost van~de Weijer}.} \bibinfo{year}{2016}\natexlab{}.
\newblock \showarticletitle{Does Multimodality Help Human and Machine for Translation and Image Captioning?}. In \bibinfo{booktitle}{\emph{Proceedings of the First Conference on Machine Translation: Volume 2, Shared Task Papers}}, \bibfield{editor}{\bibinfo{person}{Ond{\v{r}}ej Bojar}, \bibinfo{person}{Christian Buck}, \bibinfo{person}{Rajen Chatterjee}, \bibinfo{person}{Christian Federmann}, \bibinfo{person}{Liane Guillou}, \bibinfo{person}{Barry Haddow}, \bibinfo{person}{Matthias Huck}, \bibinfo{person}{Antonio~Jimeno Yepes}, \bibinfo{person}{Aur{\'e}lie N{\'e}v{\'e}ol}, \bibinfo{person}{Mariana Neves}, \bibinfo{person}{Pavel Pecina}, \bibinfo{person}{Martin Popel}, \bibinfo{person}{Philipp Koehn}, \bibinfo{person}{Christof Monz}, \bibinfo{person}{Matteo Negri}, \bibinfo{person}{Matt Post}, \bibinfo{person}{Lucia Specia}, \bibinfo{person}{Karin Verspoor}, \bibinfo{person}{J{\"o}rg Tiedemann}, {and} \bibinfo{person}{Marco Turchi}} (Eds.). \bibinfo{publisher}{Association for Computational Linguistics},
  \bibinfo{address}{Berlin, Germany}, \bibinfo{pages}{627--633}.
\newblock
\urldef\tempurl%
\url{https://doi.org/10.18653/v1/W16-2358}
\showDOI{\tempurl}


\bibitem[Caglayan et~al\mbox{.}(2019)]%
        {caglayan-etal-2019-probing}
\bibfield{author}{\bibinfo{person}{Ozan Caglayan}, \bibinfo{person}{Pranava Madhyastha}, \bibinfo{person}{Lucia Specia}, {and} \bibinfo{person}{Lo{\"i}c Barrault}.} \bibinfo{year}{2019}\natexlab{}.
\newblock \showarticletitle{Probing the Need for Visual Context in Multimodal Machine Translation}. In \bibinfo{booktitle}{\emph{Proceedings of the 2019 Conference of the North {A}merican Chapter of the Association for Computational Linguistics: Human Language Technologies, Volume 1 (Long and Short Papers)}}, \bibfield{editor}{\bibinfo{person}{Jill Burstein}, \bibinfo{person}{Christy Doran}, {and} \bibinfo{person}{Thamar Solorio}} (Eds.). \bibinfo{publisher}{Association for Computational Linguistics}, \bibinfo{address}{Minneapolis, Minnesota}, \bibinfo{pages}{4159--4170}.
\newblock
\urldef\tempurl%
\url{https://doi.org/10.18653/v1/N19-1422}
\showDOI{\tempurl}


\bibitem[Calixto and Liu(2017)]%
        {calixto-liu-2017-incorporating}
\bibfield{author}{\bibinfo{person}{Iacer Calixto} {and} \bibinfo{person}{Qun Liu}.} \bibinfo{year}{2017}\natexlab{}.
\newblock \showarticletitle{Incorporating Global Visual Features into Attention-based Neural Machine Translation.}. In \bibinfo{booktitle}{\emph{Proceedings of the 2017 Conference on Empirical Methods in Natural Language Processing}}, \bibfield{editor}{\bibinfo{person}{Martha Palmer}, \bibinfo{person}{Rebecca Hwa}, {and} \bibinfo{person}{Sebastian Riedel}} (Eds.). \bibinfo{publisher}{Association for Computational Linguistics}, \bibinfo{address}{Copenhagen, Denmark}, \bibinfo{pages}{992--1003}.
\newblock
\urldef\tempurl%
\url{https://doi.org/10.18653/v1/D17-1105}
\showDOI{\tempurl}


\bibitem[Calixto et~al\mbox{.}(2017)]%
        {calixto-etal-2017-doubly}
\bibfield{author}{\bibinfo{person}{Iacer Calixto}, \bibinfo{person}{Qun Liu}, {and} \bibinfo{person}{Nick Campbell}.} \bibinfo{year}{2017}\natexlab{}.
\newblock \showarticletitle{Doubly-Attentive Decoder for Multi-modal Neural Machine Translation}. In \bibinfo{booktitle}{\emph{Proceedings of the 55th Annual Meeting of the Association for Computational Linguistics (Volume 1: Long Papers)}}, \bibfield{editor}{\bibinfo{person}{Regina Barzilay} {and} \bibinfo{person}{Min-Yen Kan}} (Eds.). \bibinfo{publisher}{Association for Computational Linguistics}, \bibinfo{address}{Vancouver, Canada}, \bibinfo{pages}{1913--1924}.
\newblock
\urldef\tempurl%
\url{https://doi.org/10.18653/v1/P17-1175}
\showDOI{\tempurl}


\bibitem[Calixto et~al\mbox{.}(2019)]%
        {calixto-etal-2019-latent}
\bibfield{author}{\bibinfo{person}{Iacer Calixto}, \bibinfo{person}{Miguel Rios}, {and} \bibinfo{person}{Wilker Aziz}.} \bibinfo{year}{2019}\natexlab{}.
\newblock \showarticletitle{Latent Variable Model for Multi-modal Translation}. In \bibinfo{booktitle}{\emph{Proceedings of the 57th Annual Meeting of the Association for Computational Linguistics}}, \bibfield{editor}{\bibinfo{person}{Anna Korhonen}, \bibinfo{person}{David Traum}, {and} \bibinfo{person}{Llu{\'i}s M{\`a}rquez}} (Eds.). \bibinfo{publisher}{Association for Computational Linguistics}, \bibinfo{address}{Florence, Italy}, \bibinfo{pages}{6392--6405}.
\newblock
\urldef\tempurl%
\url{https://doi.org/10.18653/v1/P19-1642}
\showDOI{\tempurl}


\bibitem[Chakravarthi et~al\mbox{.}(2019)]%
        {chakravarthi-etal-2019-multilingual}
\bibfield{author}{\bibinfo{person}{Bharathi~Raja Chakravarthi}, \bibinfo{person}{Ruba Priyadharshini}, \bibinfo{person}{Bernardo Stearns}, \bibinfo{person}{Arun Jayapal}, \bibinfo{person}{Sridevy S}, \bibinfo{person}{Mihael Arcan}, \bibinfo{person}{Manel Zarrouk}, {and} \bibinfo{person}{John~P McCrae}.} \bibinfo{year}{2019}\natexlab{}.
\newblock \showarticletitle{Multilingual Multimodal Machine Translation for {D}ravidian Languages utilizing Phonetic Transcription}. In \bibinfo{booktitle}{\emph{Proceedings of the 2nd Workshop on Technologies for MT of Low Resource Languages}}, \bibfield{editor}{\bibinfo{person}{Alina Karakanta}, \bibinfo{person}{Atul~Kr. Ojha}, \bibinfo{person}{Chao-Hong Liu}, \bibinfo{person}{Jonathan Washington}, \bibinfo{person}{Nathaniel Oco}, \bibinfo{person}{Surafel~Melaku Lakew}, \bibinfo{person}{Valentin Malykh}, {and} \bibinfo{person}{Xiaobing Zhao}} (Eds.). \bibinfo{publisher}{European Association for Machine Translation}, \bibinfo{address}{Dublin, Ireland}, \bibinfo{pages}{56--63}.
\newblock
\urldef\tempurl%
\url{https://aclanthology.org/W19-6809/}
\showURL{%
\tempurl}


\bibitem[Cheng et~al\mbox{.}(2019)]%
        {cheng-etal-2019-robust}
\bibfield{author}{\bibinfo{person}{Yong Cheng}, \bibinfo{person}{Lu Jiang}, {and} \bibinfo{person}{Wolfgang Macherey}.} \bibinfo{year}{2019}\natexlab{}.
\newblock \showarticletitle{Robust Neural Machine Translation with Doubly Adversarial Inputs}. In \bibinfo{booktitle}{\emph{Proceedings of the 57th Annual Meeting of the Association for Computational Linguistics}}, \bibfield{editor}{\bibinfo{person}{Anna Korhonen}, \bibinfo{person}{David Traum}, {and} \bibinfo{person}{Llu{\'i}s M{\`a}rquez}} (Eds.). \bibinfo{publisher}{Association for Computational Linguistics}, \bibinfo{address}{Florence, Italy}, \bibinfo{pages}{4324--4333}.
\newblock
\urldef\tempurl%
\url{https://doi.org/10.18653/v1/P19-1425}
\showDOI{\tempurl}


\bibitem[Cheng et~al\mbox{.}(2018)]%
        {cheng-etal-2018-towards}
\bibfield{author}{\bibinfo{person}{Yong Cheng}, \bibinfo{person}{Zhaopeng Tu}, \bibinfo{person}{Fandong Meng}, \bibinfo{person}{Junjie Zhai}, {and} \bibinfo{person}{Yang Liu}.} \bibinfo{year}{2018}\natexlab{}.
\newblock \showarticletitle{Towards Robust Neural Machine Translation}. In \bibinfo{booktitle}{\emph{Proceedings of the 56th Annual Meeting of the Association for Computational Linguistics (Volume 1: Long Papers)}}, \bibfield{editor}{\bibinfo{person}{Iryna Gurevych} {and} \bibinfo{person}{Yusuke Miyao}} (Eds.). \bibinfo{publisher}{Association for Computational Linguistics}, \bibinfo{address}{Melbourne, Australia}, \bibinfo{pages}{1756--1766}.
\newblock
\urldef\tempurl%
\url{https://doi.org/10.18653/v1/P18-1163}
\showDOI{\tempurl}


\bibitem[Comrie(2009)]%
        {comrie2009world}
\bibfield{author}{\bibinfo{person}{Bernard Comrie}.} \bibinfo{year}{2009}\natexlab{}.
\newblock \bibinfo{booktitle}{\emph{The world's major languages}}.
\newblock \bibinfo{publisher}{Routledge}.
\newblock


\bibitem[Dabre and Sumita(2019)]%
        {dabre-sumita-2019-nicts-supervised}
\bibfield{author}{\bibinfo{person}{Raj Dabre} {and} \bibinfo{person}{Eiichiro Sumita}.} \bibinfo{year}{2019}\natexlab{}.
\newblock \showarticletitle{{NICT}`s Supervised Neural Machine Translation Systems for the {WMT}19 Translation Robustness Task}. In \bibinfo{booktitle}{\emph{Proceedings of the Fourth Conference on Machine Translation (Volume 2: Shared Task Papers, Day 1)}}, \bibfield{editor}{\bibinfo{person}{Ond{\v{r}}ej Bojar}, \bibinfo{person}{Rajen Chatterjee}, \bibinfo{person}{Christian Federmann}, \bibinfo{person}{Mark Fishel}, \bibinfo{person}{Yvette Graham}, \bibinfo{person}{Barry Haddow}, \bibinfo{person}{Matthias Huck}, \bibinfo{person}{Antonio~Jimeno Yepes}, \bibinfo{person}{Philipp Koehn}, \bibinfo{person}{Andr{\'e} Martins}, \bibinfo{person}{Christof Monz}, \bibinfo{person}{Matteo Negri}, \bibinfo{person}{Aur{\'e}lie N{\'e}v{\'e}ol}, \bibinfo{person}{Mariana Neves}, \bibinfo{person}{Matt Post}, \bibinfo{person}{Marco Turchi}, {and} \bibinfo{person}{Karin Verspoor}} (Eds.). \bibinfo{publisher}{Association for Computational Linguistics}, \bibinfo{address}{Florence, Italy}, \bibinfo{pages}{533--536}.
\newblock
\urldef\tempurl%
\url{https://doi.org/10.18653/v1/W19-5362}
\showDOI{\tempurl}


\bibitem[Dutta~Chowdhury et~al\mbox{.}(2018)]%
        {dutta-chowdhury-etal-2018-multimodal}
\bibfield{author}{\bibinfo{person}{Koel Dutta~Chowdhury}, \bibinfo{person}{Mohammed Hasanuzzaman}, {and} \bibinfo{person}{Qun Liu}.} \bibinfo{year}{2018}\natexlab{}.
\newblock \showarticletitle{Multimodal Neural Machine Translation for Low-resource Language Pairs using Synthetic Data}. In \bibinfo{booktitle}{\emph{Proceedings of the Workshop on Deep Learning Approaches for Low-Resource {NLP}}}, \bibfield{editor}{\bibinfo{person}{Reza Haffari}, \bibinfo{person}{Colin Cherry}, \bibinfo{person}{George Foster}, \bibinfo{person}{Shahram Khadivi}, {and} \bibinfo{person}{Bahar Salehi}} (Eds.). \bibinfo{publisher}{Association for Computational Linguistics}, \bibinfo{address}{Melbourne}, \bibinfo{pages}{33--42}.
\newblock
\urldef\tempurl%
\url{https://doi.org/10.18653/v1/W18-3405}
\showDOI{\tempurl}


\bibitem[Elliott et~al\mbox{.}(2016)]%
        {elliott-etal-2016-multi30k}
\bibfield{author}{\bibinfo{person}{Desmond Elliott}, \bibinfo{person}{Stella Frank}, \bibinfo{person}{Khalil Sima{'}an}, {and} \bibinfo{person}{Lucia Specia}.} \bibinfo{year}{2016}\natexlab{}.
\newblock \showarticletitle{{M}ulti30{K}: Multilingual {E}nglish-{G}erman Image Descriptions}. In \bibinfo{booktitle}{\emph{Proceedings of the 5th Workshop on Vision and Language}}, \bibfield{editor}{\bibinfo{person}{Anya Belz}, \bibinfo{person}{Erkut Erdem}, \bibinfo{person}{Krystian Mikolajczyk}, {and} \bibinfo{person}{Katerina Pastra}} (Eds.). \bibinfo{publisher}{Association for Computational Linguistics}, \bibinfo{address}{Berlin, Germany}, \bibinfo{pages}{70--74}.
\newblock
\urldef\tempurl%
\url{https://doi.org/10.18653/v1/W16-3210}
\showDOI{\tempurl}


\bibitem[Elliott and K{\'a}d{\'a}r(2017)]%
        {elliott-kadar-2017-imagination}
\bibfield{author}{\bibinfo{person}{Desmond Elliott} {and} \bibinfo{person}{{\'A}kos K{\'a}d{\'a}r}.} \bibinfo{year}{2017}\natexlab{}.
\newblock \showarticletitle{Imagination Improves Multimodal Translation}. In \bibinfo{booktitle}{\emph{Proceedings of the Eighth International Joint Conference on Natural Language Processing (Volume 1: Long Papers)}}, \bibfield{editor}{\bibinfo{person}{Greg Kondrak} {and} \bibinfo{person}{Taro Watanabe}} (Eds.). \bibinfo{publisher}{Asian Federation of Natural Language Processing}, \bibinfo{address}{Taipei, Taiwan}, \bibinfo{pages}{130--141}.
\newblock
\urldef\tempurl%
\url{https://aclanthology.org/I17-1014/}
\showURL{%
\tempurl}


\bibitem[Fernandes et~al\mbox{.}(2023)]%
        {fernandes-etal-2023-translation}
\bibfield{author}{\bibinfo{person}{Patrick Fernandes}, \bibinfo{person}{Kayo Yin}, \bibinfo{person}{Emmy Liu}, \bibinfo{person}{Andr{\'e} Martins}, {and} \bibinfo{person}{Graham Neubig}.} \bibinfo{year}{2023}\natexlab{}.
\newblock \showarticletitle{When Does Translation Require Context? A Data-driven, Multilingual Exploration}. In \bibinfo{booktitle}{\emph{Proceedings of the 61st Annual Meeting of the Association for Computational Linguistics (Volume 1: Long Papers)}}, \bibfield{editor}{\bibinfo{person}{Anna Rogers}, \bibinfo{person}{Jordan Boyd-Graber}, {and} \bibinfo{person}{Naoaki Okazaki}} (Eds.). \bibinfo{publisher}{Association for Computational Linguistics}, \bibinfo{address}{Toronto, Canada}, \bibinfo{pages}{606--626}.
\newblock
\urldef\tempurl%
\url{https://doi.org/10.18653/v1/2023.acl-long.36}
\showDOI{\tempurl}


\bibitem[Gain et~al\mbox{.}(2022a)]%
        {gain-etal-2022-investigating}
\bibfield{author}{\bibinfo{person}{Baban Gain}, \bibinfo{person}{Ramakrishna Appicharla}, \bibinfo{person}{Soumya Chennabasavaraj}, \bibinfo{person}{Nikesh Garera}, \bibinfo{person}{Asif Ekbal}, {and} \bibinfo{person}{Muthusamy Chelliah}.} \bibinfo{year}{2022}\natexlab{a}.
\newblock \showarticletitle{Investigating Effectiveness of Multi-Encoder for Conversational Neural Machine Translation}. In \bibinfo{booktitle}{\emph{Proceedings of the Seventh Conference on Machine Translation (WMT)}}, \bibfield{editor}{\bibinfo{person}{Philipp Koehn}, \bibinfo{person}{Lo{\"i}c Barrault}, \bibinfo{person}{Ond{\v{r}}ej Bojar}, \bibinfo{person}{Fethi Bougares}, \bibinfo{person}{Rajen Chatterjee}, \bibinfo{person}{Marta~R. Costa-juss{\`a}}, \bibinfo{person}{Christian Federmann}, \bibinfo{person}{Mark Fishel}, \bibinfo{person}{Alexander Fraser}, \bibinfo{person}{Markus Freitag}, \bibinfo{person}{Yvette Graham}, \bibinfo{person}{Roman Grundkiewicz}, \bibinfo{person}{Paco Guzman}, \bibinfo{person}{Barry Haddow}, \bibinfo{person}{Matthias Huck}, \bibinfo{person}{Antonio Jimeno~Yepes}, \bibinfo{person}{Tom Kocmi}, \bibinfo{person}{Andr{\'e} Martins}, \bibinfo{person}{Makoto Morishita}, \bibinfo{person}{Christof Monz}, \bibinfo{person}{Masaaki Nagata}, \bibinfo{person}{Toshiaki Nakazawa},
  \bibinfo{person}{Matteo Negri}, \bibinfo{person}{Aur{\'e}lie N{\'e}v{\'e}ol}, \bibinfo{person}{Mariana Neves}, \bibinfo{person}{Martin Popel}, \bibinfo{person}{Marco Turchi}, {and} \bibinfo{person}{Marcos Zampieri}} (Eds.). \bibinfo{publisher}{Association for Computational Linguistics}, \bibinfo{address}{Abu Dhabi, United Arab Emirates (Hybrid)}, \bibinfo{pages}{949--954}.
\newblock
\urldef\tempurl%
\url{https://aclanthology.org/2022.wmt-1.90/}
\showURL{%
\tempurl}


\bibitem[Gain et~al\mbox{.}(2022b)]%
        {gain-etal-2022-low}
\bibfield{author}{\bibinfo{person}{Baban Gain}, \bibinfo{person}{Ramakrishna Appicharla}, \bibinfo{person}{Soumya Chennabasavaraj}, \bibinfo{person}{Nikesh Garera}, \bibinfo{person}{Asif Ekbal}, {and} \bibinfo{person}{Muthusamy Chelliah}.} \bibinfo{year}{2022}\natexlab{b}.
\newblock \showarticletitle{Low Resource Chat Translation: A Benchmark for {H}indi{--}{E}nglish Language Pair}. In \bibinfo{booktitle}{\emph{Proceedings of the 15th biennial conference of the Association for Machine Translation in the Americas (Volume 1: Research Track)}}, \bibfield{editor}{\bibinfo{person}{Kevin Duh} {and} \bibinfo{person}{Francisco Guzm{\'a}n}} (Eds.). \bibinfo{publisher}{Association for Machine Translation in the Americas}, \bibinfo{address}{Orlando, USA}, \bibinfo{pages}{83--96}.
\newblock
\urldef\tempurl%
\url{https://aclanthology.org/2022.amta-research.7/}
\showURL{%
\tempurl}


\bibitem[Gain et~al\mbox{.}(2021a)]%
        {gain-etal-2021-experiences}
\bibfield{author}{\bibinfo{person}{Baban Gain}, \bibinfo{person}{Dibyanayan Bandyopadhyay}, {and} \bibinfo{person}{Asif Ekbal}.} \bibinfo{year}{2021}\natexlab{a}.
\newblock \showarticletitle{Experiences of Adapting Multimodal Machine Translation Techniques for {H}indi}. In \bibinfo{booktitle}{\emph{Proceedings of the First Workshop on Multimodal Machine Translation for Low Resource Languages (MMTLRL 2021)}}, \bibfield{editor}{\bibinfo{person}{Thoudam Doren~Singh}, \bibinfo{person}{Cristina Espa{\~n}a~i Bonet}, \bibinfo{person}{Sivaji Bandyopadhyay}, {and} \bibinfo{person}{Josef van Genabith}} (Eds.). \bibinfo{publisher}{INCOMA Ltd.}, \bibinfo{address}{Online (Virtual Mode)}, \bibinfo{pages}{40--44}.
\newblock
\urldef\tempurl%
\url{https://aclanthology.org/2021.mmtlrl-1.7/}
\showURL{%
\tempurl}


\bibitem[Gain et~al\mbox{.}(2021b)]%
        {gain-etal-2021-iitp}
\bibfield{author}{\bibinfo{person}{Baban Gain}, \bibinfo{person}{Dibyanayan Bandyopadhyay}, {and} \bibinfo{person}{Asif Ekbal}.} \bibinfo{year}{2021}\natexlab{b}.
\newblock \showarticletitle{{IITP} at {WAT} 2021: System description for {E}nglish-{H}indi Multimodal Translation Task}. In \bibinfo{booktitle}{\emph{Proceedings of the 8th Workshop on Asian Translation (WAT2021)}}, \bibfield{editor}{\bibinfo{person}{Toshiaki Nakazawa}, \bibinfo{person}{Hideki Nakayama}, \bibinfo{person}{Isao Goto}, \bibinfo{person}{Hideya Mino}, \bibinfo{person}{Chenchen Ding}, \bibinfo{person}{Raj Dabre}, \bibinfo{person}{Anoop Kunchukuttan}, \bibinfo{person}{Shohei Higashiyama}, \bibinfo{person}{Hiroshi Manabe}, \bibinfo{person}{Win~Pa Pa}, \bibinfo{person}{Shantipriya Parida}, \bibinfo{person}{Ond{\v{r}}ej Bojar}, \bibinfo{person}{Chenhui Chu}, \bibinfo{person}{Akiko Eriguchi}, \bibinfo{person}{Kaori Abe}, \bibinfo{person}{Yusuke Oda}, \bibinfo{person}{Katsuhito Sudoh}, \bibinfo{person}{Sadao Kurohashi}, {and} \bibinfo{person}{Pushpak Bhattacharyya}} (Eds.). \bibinfo{publisher}{Association for Computational Linguistics}, \bibinfo{address}{Online}, \bibinfo{pages}{161--165}.
\newblock
\urldef\tempurl%
\url{https://doi.org/10.18653/v1/2021.wat-1.18}
\showDOI{\tempurl}


\bibitem[Gain et~al\mbox{.}(2021c)]%
        {gain-et-al-not-all}
\bibfield{author}{\bibinfo{person}{Baban Gain}, \bibinfo{person}{Rejwanul Haque}, {and} \bibinfo{person}{Asif Ekbal}.} \bibinfo{year}{2021}\natexlab{c}.
\newblock \showarticletitle{Not All Contexts are Important: The Impact of Effective Context in Conversational Neural Machine Translation}. In \bibinfo{booktitle}{\emph{2021 International Joint Conference on Neural Networks (IJCNN)}}. \bibinfo{pages}{1--8}.
\newblock
\urldef\tempurl%
\url{https://doi.org/10.1109/IJCNN52387.2021.9534444}
\showDOI{\tempurl}


\bibitem[Gr{\"o}nroos et~al\mbox{.}(2018)]%
        {gronroos-etal-2018-memad}
\bibfield{author}{\bibinfo{person}{Stig-Arne Gr{\"o}nroos}, \bibinfo{person}{Benoit Huet}, \bibinfo{person}{Mikko Kurimo}, \bibinfo{person}{Jorma Laaksonen}, \bibinfo{person}{Bernard Merialdo}, \bibinfo{person}{Phu Pham}, \bibinfo{person}{Mats Sj{\"o}berg}, \bibinfo{person}{Umut Sulubacak}, \bibinfo{person}{J{\"o}rg Tiedemann}, \bibinfo{person}{Raphael Troncy}, {and} \bibinfo{person}{Ra{\'u}l V{\'a}zquez}.} \bibinfo{year}{2018}\natexlab{}.
\newblock \showarticletitle{The {M}e{MAD} Submission to the {WMT}18 Multimodal Translation Task}. In \bibinfo{booktitle}{\emph{Proceedings of the Third Conference on Machine Translation: Shared Task Papers}}, \bibfield{editor}{\bibinfo{person}{Ond{\v{r}}ej Bojar}, \bibinfo{person}{Rajen Chatterjee}, \bibinfo{person}{Christian Federmann}, \bibinfo{person}{Mark Fishel}, \bibinfo{person}{Yvette Graham}, \bibinfo{person}{Barry Haddow}, \bibinfo{person}{Matthias Huck}, \bibinfo{person}{Antonio~Jimeno Yepes}, \bibinfo{person}{Philipp Koehn}, \bibinfo{person}{Christof Monz}, \bibinfo{person}{Matteo Negri}, \bibinfo{person}{Aur{\'e}lie N{\'e}v{\'e}ol}, \bibinfo{person}{Mariana Neves}, \bibinfo{person}{Matt Post}, \bibinfo{person}{Lucia Specia}, \bibinfo{person}{Marco Turchi}, {and} \bibinfo{person}{Karin Verspoor}} (Eds.). \bibinfo{publisher}{Association for Computational Linguistics}, \bibinfo{address}{Belgium, Brussels}, \bibinfo{pages}{603--611}.
\newblock
\urldef\tempurl%
\url{https://doi.org/10.18653/v1/W18-6439}
\showDOI{\tempurl}


\bibitem[Gupta et~al\mbox{.}(2021)]%
        {gupta-etal-2021-vita}
\bibfield{author}{\bibinfo{person}{Kshitij Gupta}, \bibinfo{person}{Devansh Gautam}, {and} \bibinfo{person}{Radhika Mamidi}.} \bibinfo{year}{2021}\natexlab{}.
\newblock \showarticletitle{{V}i{TA}: Visual-Linguistic Translation by Aligning Object Tags}. In \bibinfo{booktitle}{\emph{Proceedings of the 8th Workshop on Asian Translation (WAT2021)}}, \bibfield{editor}{\bibinfo{person}{Toshiaki Nakazawa}, \bibinfo{person}{Hideki Nakayama}, \bibinfo{person}{Isao Goto}, \bibinfo{person}{Hideya Mino}, \bibinfo{person}{Chenchen Ding}, \bibinfo{person}{Raj Dabre}, \bibinfo{person}{Anoop Kunchukuttan}, \bibinfo{person}{Shohei Higashiyama}, \bibinfo{person}{Hiroshi Manabe}, \bibinfo{person}{Win~Pa Pa}, \bibinfo{person}{Shantipriya Parida}, \bibinfo{person}{Ond{\v{r}}ej Bojar}, \bibinfo{person}{Chenhui Chu}, \bibinfo{person}{Akiko Eriguchi}, \bibinfo{person}{Kaori Abe}, \bibinfo{person}{Yusuke Oda}, \bibinfo{person}{Katsuhito Sudoh}, \bibinfo{person}{Sadao Kurohashi}, {and} \bibinfo{person}{Pushpak Bhattacharyya}} (Eds.). \bibinfo{publisher}{Association for Computational Linguistics}, \bibinfo{address}{Online}, \bibinfo{pages}{166--173}.
\newblock
\urldef\tempurl%
\url{https://doi.org/10.18653/v1/2021.wat-1.19}
\showDOI{\tempurl}


\bibitem[He et~al\mbox{.}(2015)]%
        {he2015deep}
\bibfield{author}{\bibinfo{person}{Kaiming He}, \bibinfo{person}{Xiangyu Zhang}, \bibinfo{person}{Shaoqing Ren}, {and} \bibinfo{person}{Jian Sun}.} \bibinfo{year}{2015}\natexlab{}.
\newblock \bibinfo{title}{Deep Residual Learning for Image Recognition}.
\newblock
\newblock
\showeprint[arxiv]{1512.03385}~[cs.CV]


\bibitem[Helcl et~al\mbox{.}(2019)]%
        {helcl-etal-2019-cuni}
\bibfield{author}{\bibinfo{person}{Jind{\v{r}}ich Helcl}, \bibinfo{person}{Jind{\v{r}}ich Libovick{\'y}}, {and} \bibinfo{person}{Martin Popel}.} \bibinfo{year}{2019}\natexlab{}.
\newblock \showarticletitle{{CUNI} System for the {WMT}19 Robustness Task}. In \bibinfo{booktitle}{\emph{Proceedings of the Fourth Conference on Machine Translation (Volume 2: Shared Task Papers, Day 1)}}, \bibfield{editor}{\bibinfo{person}{Ond{\v{r}}ej Bojar}, \bibinfo{person}{Rajen Chatterjee}, \bibinfo{person}{Christian Federmann}, \bibinfo{person}{Mark Fishel}, \bibinfo{person}{Yvette Graham}, \bibinfo{person}{Barry Haddow}, \bibinfo{person}{Matthias Huck}, \bibinfo{person}{Antonio~Jimeno Yepes}, \bibinfo{person}{Philipp Koehn}, \bibinfo{person}{Andr{\'e} Martins}, \bibinfo{person}{Christof Monz}, \bibinfo{person}{Matteo Negri}, \bibinfo{person}{Aur{\'e}lie N{\'e}v{\'e}ol}, \bibinfo{person}{Mariana Neves}, \bibinfo{person}{Matt Post}, \bibinfo{person}{Marco Turchi}, {and} \bibinfo{person}{Karin Verspoor}} (Eds.). \bibinfo{publisher}{Association for Computational Linguistics}, \bibinfo{address}{Florence, Italy}, \bibinfo{pages}{539--543}.
\newblock
\urldef\tempurl%
\url{https://doi.org/10.18653/v1/W19-5364}
\showDOI{\tempurl}


\bibitem[Huang et~al\mbox{.}(2016)]%
        {huang-etal-2016-attention}
\bibfield{author}{\bibinfo{person}{Po-Yao Huang}, \bibinfo{person}{Frederick Liu}, \bibinfo{person}{Sz-Rung Shiang}, \bibinfo{person}{Jean Oh}, {and} \bibinfo{person}{Chris Dyer}.} \bibinfo{year}{2016}\natexlab{}.
\newblock \showarticletitle{Attention-based Multimodal Neural Machine Translation}. In \bibinfo{booktitle}{\emph{Proceedings of the First Conference on Machine Translation: Volume 2, Shared Task Papers}}, \bibfield{editor}{\bibinfo{person}{Ond{\v{r}}ej Bojar}, \bibinfo{person}{Christian Buck}, \bibinfo{person}{Rajen Chatterjee}, \bibinfo{person}{Christian Federmann}, \bibinfo{person}{Liane Guillou}, \bibinfo{person}{Barry Haddow}, \bibinfo{person}{Matthias Huck}, \bibinfo{person}{Antonio~Jimeno Yepes}, \bibinfo{person}{Aur{\'e}lie N{\'e}v{\'e}ol}, \bibinfo{person}{Mariana Neves}, \bibinfo{person}{Pavel Pecina}, \bibinfo{person}{Martin Popel}, \bibinfo{person}{Philipp Koehn}, \bibinfo{person}{Christof Monz}, \bibinfo{person}{Matteo Negri}, \bibinfo{person}{Matt Post}, \bibinfo{person}{Lucia Specia}, \bibinfo{person}{Karin Verspoor}, \bibinfo{person}{J{\"o}rg Tiedemann}, {and} \bibinfo{person}{Marco Turchi}} (Eds.). \bibinfo{publisher}{Association for Computational Linguistics}, \bibinfo{address}{Berlin,
  Germany}, \bibinfo{pages}{639--645}.
\newblock
\urldef\tempurl%
\url{https://doi.org/10.18653/v1/W16-2360}
\showDOI{\tempurl}


\bibitem[Huo et~al\mbox{.}(2020)]%
        {huo-etal-2020-diving}
\bibfield{author}{\bibinfo{person}{Jingjing Huo}, \bibinfo{person}{Christian Herold}, \bibinfo{person}{Yingbo Gao}, \bibinfo{person}{Leonard Dahlmann}, \bibinfo{person}{Shahram Khadivi}, {and} \bibinfo{person}{Hermann Ney}.} \bibinfo{year}{2020}\natexlab{}.
\newblock \showarticletitle{Diving Deep into Context-Aware Neural Machine Translation}. In \bibinfo{booktitle}{\emph{Proceedings of the Fifth Conference on Machine Translation}}, \bibfield{editor}{\bibinfo{person}{Lo{\"i}c Barrault}, \bibinfo{person}{Ond{\v{r}}ej Bojar}, \bibinfo{person}{Fethi Bougares}, \bibinfo{person}{Rajen Chatterjee}, \bibinfo{person}{Marta~R. Costa-juss{\`a}}, \bibinfo{person}{Christian Federmann}, \bibinfo{person}{Mark Fishel}, \bibinfo{person}{Alexander Fraser}, \bibinfo{person}{Yvette Graham}, \bibinfo{person}{Paco Guzman}, \bibinfo{person}{Barry Haddow}, \bibinfo{person}{Matthias Huck}, \bibinfo{person}{Antonio~Jimeno Yepes}, \bibinfo{person}{Philipp Koehn}, \bibinfo{person}{Andr{\'e} Martins}, \bibinfo{person}{Makoto Morishita}, \bibinfo{person}{Christof Monz}, \bibinfo{person}{Masaaki Nagata}, \bibinfo{person}{Toshiaki Nakazawa}, {and} \bibinfo{person}{Matteo Negri}} (Eds.). \bibinfo{publisher}{Association for Computational Linguistics}, \bibinfo{address}{Online},
  \bibinfo{pages}{604--616}.
\newblock
\urldef\tempurl%
\url{https://aclanthology.org/2020.wmt-1.71/}
\showURL{%
\tempurl}


\bibitem[Ive et~al\mbox{.}(2019)]%
        {ive-etal-2019-distilling}
\bibfield{author}{\bibinfo{person}{Julia Ive}, \bibinfo{person}{Pranava Madhyastha}, {and} \bibinfo{person}{Lucia Specia}.} \bibinfo{year}{2019}\natexlab{}.
\newblock \showarticletitle{Distilling Translations with Visual Awareness}. In \bibinfo{booktitle}{\emph{Proceedings of the 57th Annual Meeting of the Association for Computational Linguistics}}, \bibfield{editor}{\bibinfo{person}{Anna Korhonen}, \bibinfo{person}{David Traum}, {and} \bibinfo{person}{Llu{\'i}s M{\`a}rquez}} (Eds.). \bibinfo{publisher}{Association for Computational Linguistics}, \bibinfo{address}{Florence, Italy}, \bibinfo{pages}{6525--6538}.
\newblock
\urldef\tempurl%
\url{https://doi.org/10.18653/v1/P19-1653}
\showDOI{\tempurl}


\bibitem[Jin et~al\mbox{.}(2023)]%
        {jin-etal-2023-challenges}
\bibfield{author}{\bibinfo{person}{Linghao Jin}, \bibinfo{person}{Jacqueline He}, \bibinfo{person}{Jonathan May}, {and} \bibinfo{person}{Xuezhe Ma}.} \bibinfo{year}{2023}\natexlab{}.
\newblock \showarticletitle{Challenges in Context-Aware Neural Machine Translation}. In \bibinfo{booktitle}{\emph{Proceedings of the 2023 Conference on Empirical Methods in Natural Language Processing}}, \bibfield{editor}{\bibinfo{person}{Houda Bouamor}, \bibinfo{person}{Juan Pino}, {and} \bibinfo{person}{Kalika Bali}} (Eds.). \bibinfo{publisher}{Association for Computational Linguistics}, \bibinfo{address}{Singapore}, \bibinfo{pages}{15246--15263}.
\newblock
\urldef\tempurl%
\url{https://doi.org/10.18653/v1/2023.emnlp-main.943}
\showDOI{\tempurl}


\bibitem[Karpukhin et~al\mbox{.}(2019)]%
        {karpukhin-etal-2019-training}
\bibfield{author}{\bibinfo{person}{Vladimir Karpukhin}, \bibinfo{person}{Omer Levy}, \bibinfo{person}{Jacob Eisenstein}, {and} \bibinfo{person}{Marjan Ghazvininejad}.} \bibinfo{year}{2019}\natexlab{}.
\newblock \showarticletitle{Training on Synthetic Noise Improves Robustness to Natural Noise in Machine Translation}. In \bibinfo{booktitle}{\emph{Proceedings of the 5th Workshop on Noisy User-generated Text (W-NUT 2019)}}, \bibfield{editor}{\bibinfo{person}{Wei Xu}, \bibinfo{person}{Alan Ritter}, \bibinfo{person}{Tim Baldwin}, {and} \bibinfo{person}{Afshin Rahimi}} (Eds.). \bibinfo{publisher}{Association for Computational Linguistics}, \bibinfo{address}{Hong Kong, China}, \bibinfo{pages}{42--47}.
\newblock
\urldef\tempurl%
\url{https://doi.org/10.18653/v1/D19-5506}
\showDOI{\tempurl}


\bibitem[Kashani~Motlagh et~al\mbox{.}(2024)]%
        {kashani-motlagh-etal-2024-assessing}
\bibfield{author}{\bibinfo{person}{Nicholas Kashani~Motlagh}, \bibinfo{person}{Jim Davis}, \bibinfo{person}{Jeremy Gwinnup}, \bibinfo{person}{Grant Erdmann}, {and} \bibinfo{person}{Tim Anderson}.} \bibinfo{year}{2024}\natexlab{}.
\newblock \showarticletitle{Assessing the Role of Imagery in Multimodal Machine Translation}. In \bibinfo{booktitle}{\emph{Proceedings of the Ninth Conference on Machine Translation}}, \bibfield{editor}{\bibinfo{person}{Barry Haddow}, \bibinfo{person}{Tom Kocmi}, \bibinfo{person}{Philipp Koehn}, {and} \bibinfo{person}{Christof Monz}} (Eds.). \bibinfo{publisher}{Association for Computational Linguistics}, \bibinfo{address}{Miami, Florida, USA}, \bibinfo{pages}{1428--1439}.
\newblock
\urldef\tempurl%
\url{https://doi.org/10.18653/v1/2024.wmt-1.130}
\showDOI{\tempurl}


\bibitem[Khayrallah and Koehn(2018)]%
        {khayrallah-koehn-2018-impact}
\bibfield{author}{\bibinfo{person}{Huda Khayrallah} {and} \bibinfo{person}{Philipp Koehn}.} \bibinfo{year}{2018}\natexlab{}.
\newblock \showarticletitle{On the Impact of Various Types of Noise on Neural Machine Translation}. In \bibinfo{booktitle}{\emph{Proceedings of the 2nd Workshop on Neural Machine Translation and Generation}}, \bibfield{editor}{\bibinfo{person}{Alexandra Birch}, \bibinfo{person}{Andrew Finch}, \bibinfo{person}{Thang Luong}, \bibinfo{person}{Graham Neubig}, {and} \bibinfo{person}{Yusuke Oda}} (Eds.). \bibinfo{publisher}{Association for Computational Linguistics}, \bibinfo{address}{Melbourne, Australia}, \bibinfo{pages}{74--83}.
\newblock
\urldef\tempurl%
\url{https://doi.org/10.18653/v1/W18-2709}
\showDOI{\tempurl}


\bibitem[Khenglawt et~al\mbox{.}(2022)]%
        {mizo-vis-genome}
\bibfield{author}{\bibinfo{person}{Vanlalmuansangi Khenglawt}, \bibinfo{person}{Sahinur~Rahman Laskar}, \bibinfo{person}{Riyanka Manna}, \bibinfo{person}{Partha Pakray}, {and} \bibinfo{person}{Ajoy~Kumar Khan}.} \bibinfo{year}{2022}\natexlab{}.
\newblock \showarticletitle{Mizo Visual Genome 1.0 : A Dataset for English-Mizo Multimodal Neural Machine Translation}. In \bibinfo{booktitle}{\emph{2022 IEEE Silchar Subsection Conference (SILCON)}}. \bibinfo{pages}{1--6}.
\newblock
\urldef\tempurl%
\url{https://doi.org/10.1109/SILCON55242.2022.10028882}
\showDOI{\tempurl}


\bibitem[Kolesnikov et~al\mbox{.}(2021)]%
        {50650-vision-transformer}
\bibfield{author}{\bibinfo{person}{Alexander Kolesnikov}, \bibinfo{person}{Alexey Dosovitskiy}, \bibinfo{person}{Dirk Weissenborn}, \bibinfo{person}{Georg Heigold}, \bibinfo{person}{Jakob Uszkoreit}, \bibinfo{person}{Lucas Beyer}, \bibinfo{person}{Matthias Minderer}, \bibinfo{person}{Mostafa Dehghani}, \bibinfo{person}{Neil Houlsby}, \bibinfo{person}{Sylvain Gelly}, \bibinfo{person}{Thomas Unterthiner}, {and} \bibinfo{person}{Xiaohua Zhai}.} \bibinfo{year}{2021}\natexlab{}.
\newblock \showarticletitle{An Image is Worth 16x16 Words: Transformers for Image Recognition at Scale}.
\newblock


\bibitem[Kunchukuttan(2020)]%
        {kunchukuttan2020indicnlp}
\bibfield{author}{\bibinfo{person}{Anoop Kunchukuttan}.} \bibinfo{year}{2020}\natexlab{}.
\newblock \bibinfo{title}{{The IndicNLP Library}}.
\newblock \bibinfo{howpublished}{\url{https://github.com/anoopkunchukuttan/indic_nlp_library/blob/master/docs/indicnlp.pdf}}.
\newblock


\bibitem[Kunchukuttan et~al\mbox{.}(2018)]%
        {kunchukuttan-etal-2018-iit}
\bibfield{author}{\bibinfo{person}{Anoop Kunchukuttan}, \bibinfo{person}{Pratik Mehta}, {and} \bibinfo{person}{Pushpak Bhattacharyya}.} \bibinfo{year}{2018}\natexlab{}.
\newblock \showarticletitle{The {IIT} {B}ombay {E}nglish-{H}indi Parallel Corpus}. In \bibinfo{booktitle}{\emph{Proceedings of the Eleventh International Conference on Language Resources and Evaluation ({LREC} 2018)}}, \bibfield{editor}{\bibinfo{person}{Nicoletta Calzolari}, \bibinfo{person}{Khalid Choukri}, \bibinfo{person}{Christopher Cieri}, \bibinfo{person}{Thierry Declerck}, \bibinfo{person}{Sara Goggi}, \bibinfo{person}{Koiti Hasida}, \bibinfo{person}{Hitoshi Isahara}, \bibinfo{person}{Bente Maegaard}, \bibinfo{person}{Joseph Mariani}, \bibinfo{person}{H{\'e}l{\`e}ne Mazo}, \bibinfo{person}{Asuncion Moreno}, \bibinfo{person}{Jan Odijk}, \bibinfo{person}{Stelios Piperidis}, {and} \bibinfo{person}{Takenobu Tokunaga}} (Eds.). \bibinfo{publisher}{European Language Resources Association (ELRA)}, \bibinfo{address}{Miyazaki, Japan}.
\newblock
\urldef\tempurl%
\url{https://aclanthology.org/L18-1548/}
\showURL{%
\tempurl}


\bibitem[Lalrempuii et~al\mbox{.}(2021)]%
        {10.1145/3445974-mizo-multimodal-improved}
\bibfield{author}{\bibinfo{person}{Candy Lalrempuii}, \bibinfo{person}{Badal Soni}, {and} \bibinfo{person}{Partha Pakray}.} \bibinfo{year}{2021}\natexlab{}.
\newblock \showarticletitle{An Improved English-to-Mizo Neural Machine Translation}.
\newblock \bibinfo{journal}{\emph{ACM Trans. Asian Low-Resour. Lang. Inf. Process.}} \bibinfo{volume}{20}, \bibinfo{number}{4}, Article \bibinfo{articleno}{61} (\bibinfo{date}{may} \bibinfo{year}{2021}), \bibinfo{numpages}{21}~pages.
\newblock
\showISSN{2375-4699}
\urldef\tempurl%
\url{https://doi.org/10.1145/3445974}
\showDOI{\tempurl}


\bibitem[Laskar et~al\mbox{.}(2022a)]%
        {laskar-etal-2022-english}
\bibfield{author}{\bibinfo{person}{Sahinur~Rahman Laskar}, \bibinfo{person}{Pankaj Dadure}, \bibinfo{person}{Riyanka Manna}, \bibinfo{person}{Partha Pakray}, {and} \bibinfo{person}{Sivaji Bandyopadhyay}.} \bibinfo{year}{2022}\natexlab{a}.
\newblock \showarticletitle{{E}nglish to {B}engali Multimodal Neural Machine Translation using Transliteration-based Phrase Pairs Augmentation}. In \bibinfo{booktitle}{\emph{Proceedings of the 9th Workshop on Asian Translation}}. \bibinfo{publisher}{International Conference on Computational Linguistics}, \bibinfo{address}{Gyeongju, Republic of Korea}, \bibinfo{pages}{111--116}.
\newblock
\urldef\tempurl%
\url{https://aclanthology.org/2022.wat-1.14/}
\showURL{%
\tempurl}


\bibitem[Laskar et~al\mbox{.}(2021a)]%
        {laskar-etal-2021-improved}
\bibfield{author}{\bibinfo{person}{Sahinur~Rahman Laskar}, \bibinfo{person}{Abdullah Faiz Ur~Rahman Khilji}, \bibinfo{person}{Darsh Kaushik}, \bibinfo{person}{Partha Pakray}, {and} \bibinfo{person}{Sivaji Bandyopadhyay}.} \bibinfo{year}{2021}\natexlab{a}.
\newblock \showarticletitle{Improved {E}nglish to {H}indi Multimodal Neural Machine Translation}. In \bibinfo{booktitle}{\emph{Proceedings of the 8th Workshop on Asian Translation (WAT2021)}}, \bibfield{editor}{\bibinfo{person}{Toshiaki Nakazawa}, \bibinfo{person}{Hideki Nakayama}, \bibinfo{person}{Isao Goto}, \bibinfo{person}{Hideya Mino}, \bibinfo{person}{Chenchen Ding}, \bibinfo{person}{Raj Dabre}, \bibinfo{person}{Anoop Kunchukuttan}, \bibinfo{person}{Shohei Higashiyama}, \bibinfo{person}{Hiroshi Manabe}, \bibinfo{person}{Win~Pa Pa}, \bibinfo{person}{Shantipriya Parida}, \bibinfo{person}{Ond{\v{r}}ej Bojar}, \bibinfo{person}{Chenhui Chu}, \bibinfo{person}{Akiko Eriguchi}, \bibinfo{person}{Kaori Abe}, \bibinfo{person}{Yusuke Oda}, \bibinfo{person}{Katsuhito Sudoh}, \bibinfo{person}{Sadao Kurohashi}, {and} \bibinfo{person}{Pushpak Bhattacharyya}} (Eds.). \bibinfo{publisher}{Association for Computational Linguistics}, \bibinfo{address}{Online}, \bibinfo{pages}{155--160}.
\newblock
\urldef\tempurl%
\url{https://doi.org/10.18653/v1/2021.wat-1.17}
\showDOI{\tempurl}


\bibitem[Laskar et~al\mbox{.}(2020)]%
        {laskar-etal-2020-multimodal}
\bibfield{author}{\bibinfo{person}{Sahinur~Rahman Laskar}, \bibinfo{person}{Abdullah Faiz Ur~Rahman Khilji}, \bibinfo{person}{Partha Pakray}, {and} \bibinfo{person}{Sivaji Bandyopadhyay}.} \bibinfo{year}{2020}\natexlab{}.
\newblock \showarticletitle{Multimodal Neural Machine Translation for {E}nglish to {H}indi}. In \bibinfo{booktitle}{\emph{Proceedings of the 7th Workshop on Asian Translation}}, \bibfield{editor}{\bibinfo{person}{Toshiaki Nakazawa}, \bibinfo{person}{Hideki Nakayama}, \bibinfo{person}{Chenchen Ding}, \bibinfo{person}{Raj Dabre}, \bibinfo{person}{Anoop Kunchukuttan}, \bibinfo{person}{Win~Pa Pa}, \bibinfo{person}{Ond{\v{r}}ej Bojar}, \bibinfo{person}{Shantipriya Parida}, \bibinfo{person}{Isao Goto}, \bibinfo{person}{Hidaya Mino}, \bibinfo{person}{Hiroshi Manabe}, \bibinfo{person}{Katsuhito Sudoh}, \bibinfo{person}{Sadao Kurohashi}, {and} \bibinfo{person}{Pushpak Bhattacharyya}} (Eds.). \bibinfo{publisher}{Association for Computational Linguistics}, \bibinfo{address}{Suzhou, China}, \bibinfo{pages}{109--113}.
\newblock
\urldef\tempurl%
\url{https://doi.org/10.18653/v1/2020.wat-1.11}
\showDOI{\tempurl}


\bibitem[Laskar et~al\mbox{.}(2023)]%
        {LASKAR2023979-english-assamese-mmt-transliteration}
\bibfield{author}{\bibinfo{person}{Sahinur~Rahman Laskar}, \bibinfo{person}{Bishwaraj Paul}, \bibinfo{person}{Partha Pakray}, {and} \bibinfo{person}{Sivaji Bandyopadhyay}.} \bibinfo{year}{2023}\natexlab{}.
\newblock \showarticletitle{English-Assamese Multimodal Neural Machine Translation using Transliteration-based Phrase Augmentation Approach}.
\newblock \bibinfo{journal}{\emph{Procedia Computer Science}}  \bibinfo{volume}{218} (\bibinfo{year}{2023}), \bibinfo{pages}{979--988}.
\newblock
\showISSN{1877-0509}
\urldef\tempurl%
\url{https://doi.org/10.1016/j.procs.2023.01.078}
\showDOI{\tempurl}
\newblock
\shownote{International Conference on Machine Learning and Data Engineering}.


\bibitem[Laskar et~al\mbox{.}(2021b)]%
        {9752181-mmt-for-english-assaemse}
\bibfield{author}{\bibinfo{person}{Sahinur~Rahman Laskar}, \bibinfo{person}{Bishwaraj Paul}, \bibinfo{person}{Siddharth Paudwal}, \bibinfo{person}{Pranjit Gautam}, \bibinfo{person}{Nirmita Biswas}, {and} \bibinfo{person}{Partha Pakray}.} \bibinfo{year}{2021}\natexlab{b}.
\newblock \showarticletitle{Multimodal Neural Machine Translation for English–Assamese Pair}. In \bibinfo{booktitle}{\emph{2021 International Conference on Computational Performance Evaluation (ComPE)}}. \bibinfo{pages}{387--392}.
\newblock
\urldef\tempurl%
\url{https://doi.org/10.1109/ComPE53109.2021.9752181}
\showDOI{\tempurl}


\bibitem[Laskar et~al\mbox{.}(2022b)]%
        {laskar-etal-2022-investigation-english}
\bibfield{author}{\bibinfo{person}{Sahinur~Rahman Laskar}, \bibinfo{person}{Rahul Singh}, \bibinfo{person}{Md~Faizal Karim}, \bibinfo{person}{Riyanka Manna}, \bibinfo{person}{Partha Pakray}, {and} \bibinfo{person}{Sivaji Bandyopadhyay}.} \bibinfo{year}{2022}\natexlab{b}.
\newblock \showarticletitle{Investigation of {E}nglish to {H}indi Multimodal Neural Machine Translation using Transliteration-based Phrase Pairs Augmentation}. In \bibinfo{booktitle}{\emph{Proceedings of the 9th Workshop on Asian Translation}}. \bibinfo{publisher}{International Conference on Computational Linguistics}, \bibinfo{address}{Gyeongju, Republic of Korea}, \bibinfo{pages}{117--122}.
\newblock
\urldef\tempurl%
\url{https://aclanthology.org/2022.wat-1.15/}
\showURL{%
\tempurl}


\bibitem[Laskar et~al\mbox{.}(2019)]%
        {laskar-etal-2019-english}
\bibfield{author}{\bibinfo{person}{Sahinur~Rahman Laskar}, \bibinfo{person}{Rohit~Pratap Singh}, \bibinfo{person}{Partha Pakray}, {and} \bibinfo{person}{Sivaji Bandyopadhyay}.} \bibinfo{year}{2019}\natexlab{}.
\newblock \showarticletitle{{E}nglish to {H}indi Multi-modal Neural Machine Translation and {H}indi Image Captioning}. In \bibinfo{booktitle}{\emph{Proceedings of the 6th Workshop on Asian Translation}}, \bibfield{editor}{\bibinfo{person}{Toshiaki Nakazawa}, \bibinfo{person}{Chenchen Ding}, \bibinfo{person}{Raj Dabre}, \bibinfo{person}{Anoop Kunchukuttan}, \bibinfo{person}{Nobushige Doi}, \bibinfo{person}{Yusuke Oda}, \bibinfo{person}{Ond{\v{r}}ej Bojar}, \bibinfo{person}{Shantipriya Parida}, \bibinfo{person}{Isao Goto}, {and} \bibinfo{person}{Hidaya Mino}} (Eds.). \bibinfo{publisher}{Association for Computational Linguistics}, \bibinfo{address}{Hong Kong, China}, \bibinfo{pages}{62--67}.
\newblock
\urldef\tempurl%
\url{https://doi.org/10.18653/v1/D19-5205}
\showDOI{\tempurl}


\bibitem[Li et~al\mbox{.}(2020)]%
        {li-etal-2020-multi-encoder}
\bibfield{author}{\bibinfo{person}{Bei Li}, \bibinfo{person}{Hui Liu}, \bibinfo{person}{Ziyang Wang}, \bibinfo{person}{Yufan Jiang}, \bibinfo{person}{Tong Xiao}, \bibinfo{person}{Jingbo Zhu}, \bibinfo{person}{Tongran Liu}, {and} \bibinfo{person}{Changliang Li}.} \bibinfo{year}{2020}\natexlab{}.
\newblock \showarticletitle{Does Multi-Encoder Help? A Case Study on Context-Aware Neural Machine Translation}. In \bibinfo{booktitle}{\emph{Proceedings of the 58th Annual Meeting of the Association for Computational Linguistics}}, \bibfield{editor}{\bibinfo{person}{Dan Jurafsky}, \bibinfo{person}{Joyce Chai}, \bibinfo{person}{Natalie Schluter}, {and} \bibinfo{person}{Joel Tetreault}} (Eds.). \bibinfo{publisher}{Association for Computational Linguistics}, \bibinfo{address}{Online}, \bibinfo{pages}{3512--3518}.
\newblock
\urldef\tempurl%
\url{https://doi.org/10.18653/v1/2020.acl-main.322}
\showDOI{\tempurl}


\bibitem[Li et~al\mbox{.}(2022)]%
        {li-etal-2022-vision}
\bibfield{author}{\bibinfo{person}{Bei Li}, \bibinfo{person}{Chuanhao Lv}, \bibinfo{person}{Zefan Zhou}, \bibinfo{person}{Tao Zhou}, \bibinfo{person}{Tong Xiao}, \bibinfo{person}{Anxiang Ma}, {and} \bibinfo{person}{JingBo Zhu}.} \bibinfo{year}{2022}\natexlab{}.
\newblock \showarticletitle{On Vision Features in Multimodal Machine Translation}. In \bibinfo{booktitle}{\emph{Proceedings of the 60th Annual Meeting of the Association for Computational Linguistics (Volume 1: Long Papers)}}, \bibfield{editor}{\bibinfo{person}{Smaranda Muresan}, \bibinfo{person}{Preslav Nakov}, {and} \bibinfo{person}{Aline Villavicencio}} (Eds.). \bibinfo{publisher}{Association for Computational Linguistics}, \bibinfo{address}{Dublin, Ireland}, \bibinfo{pages}{6327--6337}.
\newblock
\urldef\tempurl%
\url{https://doi.org/10.18653/v1/2022.acl-long.438}
\showDOI{\tempurl}


\bibitem[Lui and Baldwin(2012)]%
        {lui-baldwin-2012-langid}
\bibfield{author}{\bibinfo{person}{Marco Lui} {and} \bibinfo{person}{Timothy Baldwin}.} \bibinfo{year}{2012}\natexlab{}.
\newblock \showarticletitle{langid.py: An Off-the-shelf Language Identification Tool}. In \bibinfo{booktitle}{\emph{Proceedings of the {ACL} 2012 System Demonstrations}}, \bibfield{editor}{\bibinfo{person}{Min Zhang}} (Ed.). \bibinfo{publisher}{Association for Computational Linguistics}, \bibinfo{address}{Jeju Island, Korea}, \bibinfo{pages}{25--30}.
\newblock
\urldef\tempurl%
\url{https://aclanthology.org/P12-3005/}
\showURL{%
\tempurl}


\bibitem[Meetei et~al\mbox{.}(2023)]%
        {MEETEI20232102-hindi-english-news}
\bibfield{author}{\bibinfo{person}{Loitongbam~Sanayai Meetei}, \bibinfo{person}{Salam~Michael Singh}, \bibinfo{person}{Alok Singh}, \bibinfo{person}{Ringki Das}, \bibinfo{person}{Thoudam~Doren Singh}, {and} \bibinfo{person}{Sivaji Bandyopadhyay}.} \bibinfo{year}{2023}\natexlab{}.
\newblock \showarticletitle{Hindi to English Multimodal Machine Translation on News Dataset in Low Resource Setting}.
\newblock \bibinfo{journal}{\emph{Procedia Computer Science}}  \bibinfo{volume}{218} (\bibinfo{year}{2023}), \bibinfo{pages}{2102--2109}.
\newblock
\showISSN{1877-0509}
\urldef\tempurl%
\url{https://doi.org/10.1016/j.procs.2023.01.186}
\showDOI{\tempurl}
\newblock
\shownote{International Conference on Machine Learning and Data Engineering}.


\bibitem[Miculicich et~al\mbox{.}(2018)]%
        {miculicich-etal-2018-document}
\bibfield{author}{\bibinfo{person}{Lesly Miculicich}, \bibinfo{person}{Dhananjay Ram}, \bibinfo{person}{Nikolaos Pappas}, {and} \bibinfo{person}{James Henderson}.} \bibinfo{year}{2018}\natexlab{}.
\newblock \showarticletitle{Document-Level Neural Machine Translation with Hierarchical Attention Networks}. In \bibinfo{booktitle}{\emph{Proceedings of the 2018 Conference on Empirical Methods in Natural Language Processing}}, \bibfield{editor}{\bibinfo{person}{Ellen Riloff}, \bibinfo{person}{David Chiang}, \bibinfo{person}{Julia Hockenmaier}, {and} \bibinfo{person}{Jun{'}ichi Tsujii}} (Eds.). \bibinfo{publisher}{Association for Computational Linguistics}, \bibinfo{address}{Brussels, Belgium}, \bibinfo{pages}{2947--2954}.
\newblock
\urldef\tempurl%
\url{https://doi.org/10.18653/v1/D18-1325}
\showDOI{\tempurl}


\bibitem[Och and Ney(2003)]%
        {och03:asc-giza-plus-plus}
\bibfield{author}{\bibinfo{person}{Franz~Josef Och} {and} \bibinfo{person}{Hermann Ney}.} \bibinfo{year}{2003}\natexlab{}.
\newblock \showarticletitle{A Systematic Comparison of Various Statistical Alignment Models}.
\newblock \bibinfo{journal}{\emph{Computational Linguistics}} \bibinfo{volume}{29}, \bibinfo{number}{1} (\bibinfo{year}{2003}), \bibinfo{pages}{19--51}.
\newblock


\bibitem[Ott et~al\mbox{.}(2019)]%
        {ott-etal-2019-fairseq}
\bibfield{author}{\bibinfo{person}{Myle Ott}, \bibinfo{person}{Sergey Edunov}, \bibinfo{person}{Alexei Baevski}, \bibinfo{person}{Angela Fan}, \bibinfo{person}{Sam Gross}, \bibinfo{person}{Nathan Ng}, \bibinfo{person}{David Grangier}, {and} \bibinfo{person}{Michael Auli}.} \bibinfo{year}{2019}\natexlab{}.
\newblock \showarticletitle{fairseq: A Fast, Extensible Toolkit for Sequence Modeling}. In \bibinfo{booktitle}{\emph{Proceedings of the 2019 Conference of the North {A}merican Chapter of the Association for Computational Linguistics (Demonstrations)}}, \bibfield{editor}{\bibinfo{person}{Waleed Ammar}, \bibinfo{person}{Annie Louis}, {and} \bibinfo{person}{Nasrin Mostafazadeh}} (Eds.). \bibinfo{publisher}{Association for Computational Linguistics}, \bibinfo{address}{Minneapolis, Minnesota}, \bibinfo{pages}{48--53}.
\newblock
\urldef\tempurl%
\url{https://doi.org/10.18653/v1/N19-4009}
\showDOI{\tempurl}


\bibitem[Papineni et~al\mbox{.}(2002)]%
        {papineni-etal-2002-bleu}
\bibfield{author}{\bibinfo{person}{Kishore Papineni}, \bibinfo{person}{Salim Roukos}, \bibinfo{person}{Todd Ward}, {and} \bibinfo{person}{Wei-Jing Zhu}.} \bibinfo{year}{2002}\natexlab{}.
\newblock \showarticletitle{{B}leu: a Method for Automatic Evaluation of Machine Translation}. In \bibinfo{booktitle}{\emph{Proceedings of the 40th Annual Meeting of the Association for Computational Linguistics}}, \bibfield{editor}{\bibinfo{person}{Pierre Isabelle}, \bibinfo{person}{Eugene Charniak}, {and} \bibinfo{person}{Dekang Lin}} (Eds.). \bibinfo{publisher}{Association for Computational Linguistics}, \bibinfo{address}{Philadelphia, Pennsylvania, USA}, \bibinfo{pages}{311--318}.
\newblock
\urldef\tempurl%
\url{https://doi.org/10.3115/1073083.1073135}
\showDOI{\tempurl}


\bibitem[Parida and Bojar(2021)]%
        {11234/1-3533-malayalam-visgen}
\bibfield{author}{\bibinfo{person}{Shantipriya Parida} {and} \bibinfo{person}{Ond{\v r}ej Bojar}.} \bibinfo{year}{2021}\natexlab{}.
\newblock \bibinfo{title}{Malayalam Visual Genome 1.0}.
\newblock
\newblock
\urldef\tempurl%
\url{http://hdl.handle.net/11234/1-3533}
\showURL{%
\tempurl}
\newblock
\shownote{{LINDAT}/{CLARIAH}-{CZ} digital library at the Institute of Formal and Applied Linguistics ({{\'U}FAL}), Faculty of Mathematics and Physics, Charles University}.


\bibitem[Parida et~al\mbox{.}(2019a)]%
        {hindi-visual-genome:2019}
\bibfield{author}{\bibinfo{person}{Shantipriya Parida}, \bibinfo{person}{Ond{\v{r}}ej Bojar}, {and} \bibinfo{person}{Satya~Ranjan Dash}.} \bibinfo{year}{2019}\natexlab{a}.
\newblock \showarticletitle{{Hindi Visual Genome: A Dataset for Multimodal English-to-Hindi Machine Translation}}.
\newblock \bibinfo{journal}{\emph{Computaci{\'o}n y Sistemas}} \bibinfo{volume}{23}, \bibinfo{number}{4} (\bibinfo{year}{2019}), \bibinfo{pages}{1499--1505}.
\newblock
\showISSN{1405-5546}
\newblock
\shownote{Presented at CICLing 2019, La Rochelle, France}.


\bibitem[Parida et~al\mbox{.}(2019b)]%
        {parida-etal-2019-idiap}
\bibfield{author}{\bibinfo{person}{Shantipriya Parida}, \bibinfo{person}{Ond{\v{r}}ej Bojar}, {and} \bibinfo{person}{Petr Motlicek}.} \bibinfo{year}{2019}\natexlab{b}.
\newblock \showarticletitle{Idiap {NMT} System for {WAT} 2019 Multimodal Translation Task}. In \bibinfo{booktitle}{\emph{Proceedings of the 6th Workshop on Asian Translation}}, \bibfield{editor}{\bibinfo{person}{Toshiaki Nakazawa}, \bibinfo{person}{Chenchen Ding}, \bibinfo{person}{Raj Dabre}, \bibinfo{person}{Anoop Kunchukuttan}, \bibinfo{person}{Nobushige Doi}, \bibinfo{person}{Yusuke Oda}, \bibinfo{person}{Ond{\v{r}}ej Bojar}, \bibinfo{person}{Shantipriya Parida}, \bibinfo{person}{Isao Goto}, {and} \bibinfo{person}{Hidaya Mino}} (Eds.). \bibinfo{publisher}{Association for Computational Linguistics}, \bibinfo{address}{Hong Kong, China}, \bibinfo{pages}{175--180}.
\newblock
\urldef\tempurl%
\url{https://doi.org/10.18653/v1/D19-5223}
\showDOI{\tempurl}


\bibitem[Parida et~al\mbox{.}(2021)]%
        {parida-etal-2021-multimodal}
\bibfield{author}{\bibinfo{person}{Shantipriya Parida}, \bibinfo{person}{Subhadarshi Panda}, \bibinfo{person}{Satya~Prakash Biswal}, \bibinfo{person}{Ketan Kotwal}, \bibinfo{person}{Arghyadeep Sen}, \bibinfo{person}{Satya~Ranjan Dash}, {and} \bibinfo{person}{Petr Motlicek}.} \bibinfo{year}{2021}\natexlab{}.
\newblock \showarticletitle{Multimodal Neural Machine Translation System for {E}nglish to {B}engali}. In \bibinfo{booktitle}{\emph{Proceedings of the First Workshop on Multimodal Machine Translation for Low Resource Languages (MMTLRL 2021)}}, \bibfield{editor}{\bibinfo{person}{Thoudam Doren~Singh}, \bibinfo{person}{Cristina Espa{\~n}a~i Bonet}, \bibinfo{person}{Sivaji Bandyopadhyay}, {and} \bibinfo{person}{Josef van Genabith}} (Eds.). \bibinfo{publisher}{INCOMA Ltd.}, \bibinfo{address}{Online (Virtual Mode)}, \bibinfo{pages}{31--39}.
\newblock
\urldef\tempurl%
\url{https://aclanthology.org/2021.mmtlrl-1.6/}
\showURL{%
\tempurl}


\bibitem[Parida et~al\mbox{.}(2022)]%
        {parida-etal-2022-silo}
\bibfield{author}{\bibinfo{person}{Shantipriya Parida}, \bibinfo{person}{Subhadarshi Panda}, \bibinfo{person}{Stig-Arne Gr{\"o}nroos}, \bibinfo{person}{Mark Granroth-Wilding}, {and} \bibinfo{person}{Mika Koistinen}.} \bibinfo{year}{2022}\natexlab{}.
\newblock \showarticletitle{Silo {NLP}`s Participation at {WAT}2022}. In \bibinfo{booktitle}{\emph{Proceedings of the 9th Workshop on Asian Translation}}. \bibinfo{publisher}{International Conference on Computational Linguistics}, \bibinfo{address}{Gyeongju, Republic of Korea}, \bibinfo{pages}{99--105}.
\newblock
\urldef\tempurl%
\url{https://aclanthology.org/2022.wat-1.12/}
\showURL{%
\tempurl}


\bibitem[Passban et~al\mbox{.}(2021)]%
        {passban-etal-2021-revisiting-robust}
\bibfield{author}{\bibinfo{person}{Peyman Passban}, \bibinfo{person}{Puneeth Saladi}, {and} \bibinfo{person}{Qun Liu}.} \bibinfo{year}{2021}\natexlab{}.
\newblock \showarticletitle{Revisiting Robust Neural Machine Translation: A Transformer Case Study}. In \bibinfo{booktitle}{\emph{Findings of the Association for Computational Linguistics: EMNLP 2021}}, \bibfield{editor}{\bibinfo{person}{Marie-Francine Moens}, \bibinfo{person}{Xuanjing Huang}, \bibinfo{person}{Lucia Specia}, {and} \bibinfo{person}{Scott Wen-tau Yih}} (Eds.). \bibinfo{publisher}{Association for Computational Linguistics}, \bibinfo{address}{Punta Cana, Dominican Republic}, \bibinfo{pages}{3831--3840}.
\newblock
\urldef\tempurl%
\url{https://doi.org/10.18653/v1/2021.findings-emnlp.323}
\showDOI{\tempurl}


\bibitem[Pathak et~al\mbox{.}(2018)]%
        {Pathak2018EnglishMizoMT-english-mizo-neural-statistical}
\bibfield{author}{\bibinfo{person}{Amarnath Pathak}, \bibinfo{person}{Partha Pakray}, {and} \bibinfo{person}{Jereemi Bentham}.} \bibinfo{year}{2018}\natexlab{}.
\newblock \showarticletitle{English–Mizo Machine Translation using neural and statistical approaches}.
\newblock \bibinfo{journal}{\emph{Neural Computing and Applications}} (\bibinfo{year}{2018}), \bibinfo{pages}{1--17}.
\newblock


\bibitem[Peng et~al\mbox{.}(2022)]%
        {peng-etal-2022-distill}
\bibfield{author}{\bibinfo{person}{Ru Peng}, \bibinfo{person}{Yawen Zeng}, {and} \bibinfo{person}{Jake Zhao}.} \bibinfo{year}{2022}\natexlab{}.
\newblock \showarticletitle{Distill The Image to Nowhere: Inversion Knowledge Distillation for Multimodal Machine Translation}. In \bibinfo{booktitle}{\emph{Proceedings of the 2022 Conference on Empirical Methods in Natural Language Processing}}, \bibfield{editor}{\bibinfo{person}{Yoav Goldberg}, \bibinfo{person}{Zornitsa Kozareva}, {and} \bibinfo{person}{Yue Zhang}} (Eds.). \bibinfo{publisher}{Association for Computational Linguistics}, \bibinfo{address}{Abu Dhabi, United Arab Emirates}, \bibinfo{pages}{2379--2390}.
\newblock
\urldef\tempurl%
\url{https://doi.org/10.18653/v1/2022.emnlp-main.152}
\showDOI{\tempurl}


\bibitem[Popovi{\'c}(2015)]%
        {popovic-2015-chrf}
\bibfield{author}{\bibinfo{person}{Maja Popovi{\'c}}.} \bibinfo{year}{2015}\natexlab{}.
\newblock \showarticletitle{chr{F}: character n-gram {F}-score for automatic {MT} evaluation}. In \bibinfo{booktitle}{\emph{Proceedings of the Tenth Workshop on Statistical Machine Translation}}, \bibfield{editor}{\bibinfo{person}{Ond{\v{r}}ej Bojar}, \bibinfo{person}{Rajan Chatterjee}, \bibinfo{person}{Christian Federmann}, \bibinfo{person}{Barry Haddow}, \bibinfo{person}{Chris Hokamp}, \bibinfo{person}{Matthias Huck}, \bibinfo{person}{Varvara Logacheva}, {and} \bibinfo{person}{Pavel Pecina}} (Eds.). \bibinfo{publisher}{Association for Computational Linguistics}, \bibinfo{address}{Lisbon, Portugal}, \bibinfo{pages}{392--395}.
\newblock
\urldef\tempurl%
\url{https://doi.org/10.18653/v1/W15-3049}
\showDOI{\tempurl}


\bibitem[Popovi{\'c}(2016)]%
        {popovic-2016-chrf}
\bibfield{author}{\bibinfo{person}{Maja Popovi{\'c}}.} \bibinfo{year}{2016}\natexlab{}.
\newblock \showarticletitle{chr{F} deconstructed: beta parameters and n-gram weights}. In \bibinfo{booktitle}{\emph{Proceedings of the First Conference on Machine Translation: Volume 2, Shared Task Papers}}, \bibfield{editor}{\bibinfo{person}{Ond{\v{r}}ej Bojar}, \bibinfo{person}{Christian Buck}, \bibinfo{person}{Rajen Chatterjee}, \bibinfo{person}{Christian Federmann}, \bibinfo{person}{Liane Guillou}, \bibinfo{person}{Barry Haddow}, \bibinfo{person}{Matthias Huck}, \bibinfo{person}{Antonio~Jimeno Yepes}, \bibinfo{person}{Aur{\'e}lie N{\'e}v{\'e}ol}, \bibinfo{person}{Mariana Neves}, \bibinfo{person}{Pavel Pecina}, \bibinfo{person}{Martin Popel}, \bibinfo{person}{Philipp Koehn}, \bibinfo{person}{Christof Monz}, \bibinfo{person}{Matteo Negri}, \bibinfo{person}{Matt Post}, \bibinfo{person}{Lucia Specia}, \bibinfo{person}{Karin Verspoor}, \bibinfo{person}{J{\"o}rg Tiedemann}, {and} \bibinfo{person}{Marco Turchi}} (Eds.). \bibinfo{publisher}{Association for Computational Linguistics}, \bibinfo{address}{Berlin,
  Germany}, \bibinfo{pages}{499--504}.
\newblock
\urldef\tempurl%
\url{https://doi.org/10.18653/v1/W16-2341}
\showDOI{\tempurl}


\bibitem[Radford et~al\mbox{.}(2021)]%
        {radford2021learningtransferablevisualmodels}
\bibfield{author}{\bibinfo{person}{Alec Radford}, \bibinfo{person}{Jong~Wook Kim}, \bibinfo{person}{Chris Hallacy}, \bibinfo{person}{Aditya Ramesh}, \bibinfo{person}{Gabriel Goh}, \bibinfo{person}{Sandhini Agarwal}, \bibinfo{person}{Girish Sastry}, \bibinfo{person}{Amanda Askell}, \bibinfo{person}{Pamela Mishkin}, \bibinfo{person}{Jack Clark}, \bibinfo{person}{Gretchen Krueger}, {and} \bibinfo{person}{Ilya Sutskever}.} \bibinfo{year}{2021}\natexlab{}.
\newblock \bibinfo{title}{Learning Transferable Visual Models From Natural Language Supervision}.
\newblock
\newblock
\showeprint[arxiv]{2103.00020}~[cs.CV]
\urldef\tempurl%
\url{https://arxiv.org/abs/2103.00020}
\showURL{%
\tempurl}


\bibitem[Ramesh et~al\mbox{.}(2022)]%
        {ramesh-etal-2022-samanantar}
\bibfield{author}{\bibinfo{person}{Gowtham Ramesh}, \bibinfo{person}{Sumanth Doddapaneni}, \bibinfo{person}{Aravinth Bheemaraj}, \bibinfo{person}{Mayank Jobanputra}, \bibinfo{person}{Raghavan AK}, \bibinfo{person}{Ajitesh Sharma}, \bibinfo{person}{Sujit Sahoo}, \bibinfo{person}{Harshita Diddee}, \bibinfo{person}{Mahalakshmi J}, \bibinfo{person}{Divyanshu Kakwani}, \bibinfo{person}{Navneet Kumar}, \bibinfo{person}{Aswin Pradeep}, \bibinfo{person}{Srihari Nagaraj}, \bibinfo{person}{Kumar Deepak}, \bibinfo{person}{Vivek Raghavan}, \bibinfo{person}{Anoop Kunchukuttan}, \bibinfo{person}{Pratyush Kumar}, {and} \bibinfo{person}{Mitesh~Shantadevi Khapra}.} \bibinfo{year}{2022}\natexlab{}.
\newblock \showarticletitle{Samanantar: The Largest Publicly Available Parallel Corpora Collection for 11 {I}ndic Languages}.
\newblock \bibinfo{journal}{\emph{Transactions of the Association for Computational Linguistics}}  \bibinfo{volume}{10} (\bibinfo{year}{2022}), \bibinfo{pages}{145--162}.
\newblock
\urldef\tempurl%
\url{https://doi.org/10.1162/tacl_a_00452}
\showDOI{\tempurl}


\bibitem[Sanayai~Meetei et~al\mbox{.}(2019)]%
        {sanayai-meetei-etal-2019-wat2019}
\bibfield{author}{\bibinfo{person}{Loitongbam Sanayai~Meetei}, \bibinfo{person}{Thoudam~Doren Singh}, {and} \bibinfo{person}{Sivaji Bandyopadhyay}.} \bibinfo{year}{2019}\natexlab{}.
\newblock \showarticletitle{{WAT}2019: {E}nglish-{H}indi Translation on {H}indi Visual Genome Dataset}. In \bibinfo{booktitle}{\emph{Proceedings of the 6th Workshop on Asian Translation}}, \bibfield{editor}{\bibinfo{person}{Toshiaki Nakazawa}, \bibinfo{person}{Chenchen Ding}, \bibinfo{person}{Raj Dabre}, \bibinfo{person}{Anoop Kunchukuttan}, \bibinfo{person}{Nobushige Doi}, \bibinfo{person}{Yusuke Oda}, \bibinfo{person}{Ond{\v{r}}ej Bojar}, \bibinfo{person}{Shantipriya Parida}, \bibinfo{person}{Isao Goto}, {and} \bibinfo{person}{Hidaya Mino}} (Eds.). \bibinfo{publisher}{Association for Computational Linguistics}, \bibinfo{address}{Hong Kong, China}, \bibinfo{pages}{181--188}.
\newblock
\urldef\tempurl%
\url{https://doi.org/10.18653/v1/D19-5224}
\showDOI{\tempurl}


\bibitem[Sen et~al\mbox{.}(2022)]%
        {sen2022bengali-visgen}
\bibfield{author}{\bibinfo{person}{Arghyadeep Sen}, \bibinfo{person}{Shantipriya Parida}, \bibinfo{person}{Ketan Kotwal}, \bibinfo{person}{Subhadarshi Panda}, \bibinfo{person}{Ond{\v{r}}ej Bojar}, {and} \bibinfo{person}{Satya~Ranjan Dash}.} \bibinfo{year}{2022}\natexlab{}.
\newblock \showarticletitle{Bengali Visual Genome: A Multimodal Dataset for Machine Translation and Image Captioning}.
\newblock In \bibinfo{booktitle}{\emph{Intelligent Data Engineering and Analytics}}. \bibinfo{publisher}{Springer}, \bibinfo{pages}{63--70}.
\newblock


\bibitem[Shan et~al\mbox{.}(2022)]%
        {shan2022ernieunix2}
\bibfield{author}{\bibinfo{person}{Bin Shan}, \bibinfo{person}{Yaqian Han}, \bibinfo{person}{Weichong Yin}, \bibinfo{person}{Shuohuan Wang}, \bibinfo{person}{Yu Sun}, \bibinfo{person}{Hao Tian}, \bibinfo{person}{Hua Wu}, {and} \bibinfo{person}{Haifeng Wang}.} \bibinfo{year}{2022}\natexlab{}.
\newblock \bibinfo{title}{ERNIE-UniX2: A Unified Cross-lingual Cross-modal Framework for Understanding and Generation}.
\newblock
\newblock
\showeprint[arxiv]{2211.04861}~[cs.CV]


\bibitem[Shi and Yu(2022)]%
        {shi2022adding-visual-info}
\bibfield{author}{\bibinfo{person}{Xiayang Shi} {and} \bibinfo{person}{Zhenqiang Yu}.} \bibinfo{year}{2022}\natexlab{}.
\newblock \showarticletitle{Adding Visual Information to Improve Multimodal Machine Translation for Low-Resource Language}.
\newblock \bibinfo{journal}{\emph{Mathematical Problems in Engineering}}  \bibinfo{volume}{2022} (\bibinfo{year}{2022}).
\newblock


\bibitem[Simonyan and Zisserman(2014)]%
        {https://doi.org/10.48550/arxiv.1409.1556-vgg}
\bibfield{author}{\bibinfo{person}{Karen Simonyan} {and} \bibinfo{person}{Andrew Zisserman}.} \bibinfo{year}{2014}\natexlab{}.
\newblock \bibinfo{title}{Very Deep Convolutional Networks for Large-Scale Image Recognition}.
\newblock
\newblock
\urldef\tempurl%
\url{https://doi.org/10.48550/ARXIV.1409.1556}
\showDOI{\tempurl}


\bibitem[Singh et~al\mbox{.}(2021)]%
        {singh-etal-2021-multiple}
\bibfield{author}{\bibinfo{person}{Salam~Michael Singh}, \bibinfo{person}{Loitongbam Sanayai~Meetei}, \bibinfo{person}{Thoudam~Doren Singh}, {and} \bibinfo{person}{Sivaji Bandyopadhyay}.} \bibinfo{year}{2021}\natexlab{}.
\newblock \showarticletitle{Multiple Captions Embellished Multilingual Multi-Modal Neural Machine Translation}. In \bibinfo{booktitle}{\emph{Proceedings of the First Workshop on Multimodal Machine Translation for Low Resource Languages (MMTLRL 2021)}}, \bibfield{editor}{\bibinfo{person}{Thoudam Doren~Singh}, \bibinfo{person}{Cristina Espa{\~n}a~i Bonet}, \bibinfo{person}{Sivaji Bandyopadhyay}, {and} \bibinfo{person}{Josef van Genabith}} (Eds.). \bibinfo{publisher}{INCOMA Ltd.}, \bibinfo{address}{Online (Virtual Mode)}, \bibinfo{pages}{2--11}.
\newblock
\urldef\tempurl%
\url{https://aclanthology.org/2021.mmtlrl-1.2/}
\showURL{%
\tempurl}


\bibitem[Snover et~al\mbox{.}(2006)]%
        {snover-etal-2006-study}
\bibfield{author}{\bibinfo{person}{Matthew Snover}, \bibinfo{person}{Bonnie Dorr}, \bibinfo{person}{Rich Schwartz}, \bibinfo{person}{Linnea Micciulla}, {and} \bibinfo{person}{John Makhoul}.} \bibinfo{year}{2006}\natexlab{}.
\newblock \showarticletitle{A Study of Translation Edit Rate with Targeted Human Annotation}. In \bibinfo{booktitle}{\emph{Proceedings of the 7th Conference of the Association for Machine Translation in the Americas: Technical Papers}}. \bibinfo{publisher}{Association for Machine Translation in the Americas}, \bibinfo{address}{Cambridge, Massachusetts, USA}, \bibinfo{pages}{223--231}.
\newblock
\urldef\tempurl%
\url{https://aclanthology.org/2006.amta-papers.25/}
\showURL{%
\tempurl}


\bibitem[Sperber et~al\mbox{.}(2017)]%
        {sperber-etal-2017-toward}
\bibfield{author}{\bibinfo{person}{Matthias Sperber}, \bibinfo{person}{Jan Niehues}, {and} \bibinfo{person}{Alex Waibel}.} \bibinfo{year}{2017}\natexlab{}.
\newblock \showarticletitle{Toward Robust Neural Machine Translation for Noisy Input Sequences}. In \bibinfo{booktitle}{\emph{Proceedings of the 14th International Conference on Spoken Language Translation}}, \bibfield{editor}{\bibinfo{person}{Sakriani Sakti} {and} \bibinfo{person}{Masao Utiyama}} (Eds.). \bibinfo{publisher}{International Workshop on Spoken Language Translation}, \bibinfo{address}{Tokyo, Japan}, \bibinfo{pages}{90--96}.
\newblock
\urldef\tempurl%
\url{https://aclanthology.org/2017.iwslt-1.13/}
\showURL{%
\tempurl}


\bibitem[Tewel et~al\mbox{.}(2022)]%
        {9878503}
\bibfield{author}{\bibinfo{person}{Yoad Tewel}, \bibinfo{person}{Yoav Shalev}, \bibinfo{person}{Idan Schwartz}, {and} \bibinfo{person}{Lior Wolf}.} \bibinfo{year}{2022}\natexlab{}.
\newblock \showarticletitle{ZeroCap: Zero-Shot Image-to-Text Generation for Visual-Semantic Arithmetic}. In \bibinfo{booktitle}{\emph{2022 IEEE/CVF Conference on Computer Vision and Pattern Recognition (CVPR)}}. \bibinfo{pages}{17897--17907}.
\newblock
\urldef\tempurl%
\url{https://doi.org/10.1109/CVPR52688.2022.01739}
\showDOI{\tempurl}


\bibitem[Thihlum et~al\mbox{.}(2020)]%
        {9500022-mt-english-to-mizo}
\bibfield{author}{\bibinfo{person}{Zaitinkhuma Thihlum}, \bibinfo{person}{Vanlalmuansangi Khenglawt}, {and} \bibinfo{person}{Somen Debnath}.} \bibinfo{year}{2020}\natexlab{}.
\newblock \showarticletitle{Machine Translation of English Language to Mizo Language}. In \bibinfo{booktitle}{\emph{2020 IEEE International Conference on Cloud Computing in Emerging Markets (CCEM)}}. \bibinfo{pages}{92--97}.
\newblock
\urldef\tempurl%
\url{https://doi.org/10.1109/CCEM50674.2020.00028}
\showDOI{\tempurl}


\bibitem[Tiedemann and Scherrer(2017)]%
        {tiedemann-scherrer-2017-neural}
\bibfield{author}{\bibinfo{person}{J{\"o}rg Tiedemann} {and} \bibinfo{person}{Yves Scherrer}.} \bibinfo{year}{2017}\natexlab{}.
\newblock \showarticletitle{Neural Machine Translation with Extended Context}. In \bibinfo{booktitle}{\emph{Proceedings of the Third Workshop on Discourse in Machine Translation}}, \bibfield{editor}{\bibinfo{person}{Bonnie Webber}, \bibinfo{person}{Andrei Popescu-Belis}, {and} \bibinfo{person}{J{\"o}rg Tiedemann}} (Eds.). \bibinfo{publisher}{Association for Computational Linguistics}, \bibinfo{address}{Copenhagen, Denmark}, \bibinfo{pages}{82--92}.
\newblock
\urldef\tempurl%
\url{https://doi.org/10.18653/v1/W17-4811}
\showDOI{\tempurl}


\bibitem[Vaswani et~al\mbox{.}(2017a)]%
        {attention-is-all-you-need}
\bibfield{author}{\bibinfo{person}{Ashish Vaswani}, \bibinfo{person}{Noam Shazeer}, \bibinfo{person}{Niki Parmar}, \bibinfo{person}{Jakob Uszkoreit}, \bibinfo{person}{Llion Jones}, \bibinfo{person}{Aidan~N Gomez}, \bibinfo{person}{\L~ukasz Kaiser}, {and} \bibinfo{person}{Illia Polosukhin}.} \bibinfo{year}{2017}\natexlab{a}.
\newblock \showarticletitle{Attention is All you Need}. In \bibinfo{booktitle}{\emph{Advances in Neural Information Processing Systems}}, \bibfield{editor}{\bibinfo{person}{I.~Guyon}, \bibinfo{person}{U.~Von Luxburg}, \bibinfo{person}{S.~Bengio}, \bibinfo{person}{H.~Wallach}, \bibinfo{person}{R.~Fergus}, \bibinfo{person}{S.~Vishwanathan}, {and} \bibinfo{person}{R.~Garnett}} (Eds.), Vol.~\bibinfo{volume}{30}. \bibinfo{publisher}{Curran Associates, Inc.}
\newblock
\urldef\tempurl%
\url{https://proceedings.neurips.cc/paper_files/paper/2017/file/3f5ee243547dee91fbd053c1c4a845aa-Paper.pdf}
\showURL{%
\tempurl}


\bibitem[Vaswani et~al\mbox{.}(2017b)]%
        {NIPS2017_3f5ee243-vaswani-transformer}
\bibfield{author}{\bibinfo{person}{Ashish Vaswani}, \bibinfo{person}{Noam Shazeer}, \bibinfo{person}{Niki Parmar}, \bibinfo{person}{Jakob Uszkoreit}, \bibinfo{person}{Llion Jones}, \bibinfo{person}{Aidan~N Gomez}, \bibinfo{person}{\L~ukasz Kaiser}, {and} \bibinfo{person}{Illia Polosukhin}.} \bibinfo{year}{2017}\natexlab{b}.
\newblock \showarticletitle{Attention is All you Need}. In \bibinfo{booktitle}{\emph{Advances in Neural Information Processing Systems}}, \bibfield{editor}{\bibinfo{person}{I.~Guyon}, \bibinfo{person}{U.~Von Luxburg}, \bibinfo{person}{S.~Bengio}, \bibinfo{person}{H.~Wallach}, \bibinfo{person}{R.~Fergus}, \bibinfo{person}{S.~Vishwanathan}, {and} \bibinfo{person}{R.~Garnett}} (Eds.), Vol.~\bibinfo{volume}{30}. \bibinfo{publisher}{Curran Associates, Inc.}
\newblock
\urldef\tempurl%
\url{https://proceedings.neurips.cc/paper/2017/file/3f5ee243547dee91fbd053c1c4a845aa-Paper.pdf}
\showURL{%
\tempurl}


\bibitem[Wang and Ng(2013)]%
        {wang-ng-2013-beam}
\bibfield{author}{\bibinfo{person}{Pidong Wang} {and} \bibinfo{person}{Hwee~Tou Ng}.} \bibinfo{year}{2013}\natexlab{}.
\newblock \showarticletitle{A Beam-Search Decoder for Normalization of Social Media Text with Application to Machine Translation}. In \bibinfo{booktitle}{\emph{Proceedings of the 2013 Conference of the North {A}merican Chapter of the Association for Computational Linguistics: Human Language Technologies}}, \bibfield{editor}{\bibinfo{person}{Lucy Vanderwende}, \bibinfo{person}{Hal Daum{\'e}~III}, {and} \bibinfo{person}{Katrin Kirchhoff}} (Eds.). \bibinfo{publisher}{Association for Computational Linguistics}, \bibinfo{address}{Atlanta, Georgia}, \bibinfo{pages}{471--481}.
\newblock
\urldef\tempurl%
\url{https://aclanthology.org/N13-1050/}
\showURL{%
\tempurl}


\bibitem[Wang et~al\mbox{.}(2021)]%
        {wang-etal-2021-secoco-self}
\bibfield{author}{\bibinfo{person}{Tao Wang}, \bibinfo{person}{Chengqi Zhao}, \bibinfo{person}{Mingxuan Wang}, \bibinfo{person}{Lei Li}, \bibinfo{person}{Hang Li}, {and} \bibinfo{person}{Deyi Xiong}.} \bibinfo{year}{2021}\natexlab{}.
\newblock \showarticletitle{Secoco: Self-Correcting Encoding for Neural Machine Translation}. In \bibinfo{booktitle}{\emph{Findings of the Association for Computational Linguistics: EMNLP 2021}}, \bibfield{editor}{\bibinfo{person}{Marie-Francine Moens}, \bibinfo{person}{Xuanjing Huang}, \bibinfo{person}{Lucia Specia}, {and} \bibinfo{person}{Scott Wen-tau Yih}} (Eds.). \bibinfo{publisher}{Association for Computational Linguistics}, \bibinfo{address}{Punta Cana, Dominican Republic}, \bibinfo{pages}{4639--4644}.
\newblock
\urldef\tempurl%
\url{https://doi.org/10.18653/v1/2021.findings-emnlp.396}
\showDOI{\tempurl}


\bibitem[Xie et~al\mbox{.}(2017)]%
        {xie2017data-noising}
\bibfield{author}{\bibinfo{person}{Ziang Xie}, \bibinfo{person}{Sida~I. Wang}, \bibinfo{person}{Jiwei Li}, \bibinfo{person}{Daniel L{\'e}vy}, \bibinfo{person}{Aiming Nie}, \bibinfo{person}{Dan Jurafsky}, {and} \bibinfo{person}{Andrew~Y. Ng}.} \bibinfo{year}{2017}\natexlab{}.
\newblock \showarticletitle{Data Noising as Smoothing in Neural Network Language Models}. In \bibinfo{booktitle}{\emph{International Conference on Learning Representations}}.
\newblock
\urldef\tempurl%
\url{https://openreview.net/forum?id=H1VyHY9gg}
\showURL{%
\tempurl}


\bibitem[Yao and Wan(2020)]%
        {yao-wan-2020-multimodal}
\bibfield{author}{\bibinfo{person}{Shaowei Yao} {and} \bibinfo{person}{Xiaojun Wan}.} \bibinfo{year}{2020}\natexlab{}.
\newblock \showarticletitle{Multimodal Transformer for Multimodal Machine Translation}. In \bibinfo{booktitle}{\emph{Proceedings of the 58th Annual Meeting of the Association for Computational Linguistics}}, \bibfield{editor}{\bibinfo{person}{Dan Jurafsky}, \bibinfo{person}{Joyce Chai}, \bibinfo{person}{Natalie Schluter}, {and} \bibinfo{person}{Joel Tetreault}} (Eds.). \bibinfo{publisher}{Association for Computational Linguistics}, \bibinfo{address}{Online}, \bibinfo{pages}{4346--4350}.
\newblock
\urldef\tempurl%
\url{https://doi.org/10.18653/v1/2020.acl-main.400}
\showDOI{\tempurl}


\bibitem[Zhao et~al\mbox{.}(2022)]%
        {ZHAO20221-region-attentive}
\bibfield{author}{\bibinfo{person}{Yuting Zhao}, \bibinfo{person}{Mamoru Komachi}, \bibinfo{person}{Tomoyuki Kajiwara}, {and} \bibinfo{person}{Chenhui Chu}.} \bibinfo{year}{2022}\natexlab{}.
\newblock \showarticletitle{Region-attentive multimodal neural machine translation}.
\newblock \bibinfo{journal}{\emph{Neurocomputing}}  \bibinfo{volume}{476} (\bibinfo{year}{2022}), \bibinfo{pages}{1--13}.
\newblock
\showISSN{0925-2312}
\urldef\tempurl%
\url{https://doi.org/10.1016/j.neucom.2021.12.076}
\showDOI{\tempurl}


\bibitem[Zhou et~al\mbox{.}(2018)]%
        {zhou-etal-2018-visual}
\bibfield{author}{\bibinfo{person}{Mingyang Zhou}, \bibinfo{person}{Runxiang Cheng}, \bibinfo{person}{Yong~Jae Lee}, {and} \bibinfo{person}{Zhou Yu}.} \bibinfo{year}{2018}\natexlab{}.
\newblock \showarticletitle{A Visual Attention Grounding Neural Model for Multimodal Machine Translation}. In \bibinfo{booktitle}{\emph{Proceedings of the 2018 Conference on Empirical Methods in Natural Language Processing}}, \bibfield{editor}{\bibinfo{person}{Ellen Riloff}, \bibinfo{person}{David Chiang}, \bibinfo{person}{Julia Hockenmaier}, {and} \bibinfo{person}{Jun{'}ichi Tsujii}} (Eds.). \bibinfo{publisher}{Association for Computational Linguistics}, \bibinfo{address}{Brussels, Belgium}, \bibinfo{pages}{3643--3653}.
\newblock
\urldef\tempurl%
\url{https://doi.org/10.18653/v1/D18-1400}
\showDOI{\tempurl}


\end{thebibliography}

\appendix

\end{document}